\batchmode
\makeatletter
\def\input@path{{"C:/Trabajo laptop/Mis articulos/Finished/Distributed Fusion PMB/"}}
\makeatother
\documentclass[onecolumn,british,twocolumn]{IEEEtran}
\usepackage[T1]{fontenc}
\usepackage[latin9]{inputenc}
\usepackage{color}
\usepackage{float}
\usepackage{amsmath}
\usepackage{amsthm}
\usepackage{amssymb}
\usepackage{graphicx}
\PassOptionsToPackage{normalem}{ulem}
\usepackage{ulem}

\makeatletter

\providecommand{\tabularnewline}{\\}
\floatstyle{ruled}
\newfloat{algorithm}{tbp}{loa}
\providecommand{\algorithmname}{Algorithm}
\floatname{algorithm}{\protect\algorithmname}

\theoremstyle{plain}
\newtheorem{thm}{\protect\theoremname}
\theoremstyle{plain}
\newtheorem{lem}[thm]{\protect\lemmaname}
\theoremstyle{definition}
\newtheorem{example}[thm]{\protect\examplename}

\usepackage{cite} 
\usepackage[margin=8pt,font=footnotesize]{caption}
\usepackage{algorithm}
\usepackage{algpseudocode}

\usepackage{amsmath} 
\allowdisplaybreaks

\makeatother

\usepackage{babel}
\providecommand{\examplename}{Example}
\providecommand{\lemmaname}{Lemma}
\providecommand{\theoremname}{Theorem}

\begin{document}
\title{Distributed Poisson multi-Bernoulli filtering via generalised covariance
intersection}
\author{Ángel F. García-Fernández, Giorgio Battistelli\thanks{A. F. Garc\'ia-Fern\'andez is with the IPTC, ETSI de Telecomunicaci\'on, Universidad Polit\'ecnica de Madrid, 28040 Madrid, Spain (email: \mbox{angel.garcia.fernandez@upm.es}).} 
\thanks{G. Battistelli is with Dipartimento di Ingegneria dell'Informazione (DINFO), Universit\`a di Firenze, Via Santa Marta 3, 50139, Firenze, Italy (e-mail: giorgio.battistelli@unifi.it).}
\thanks{The authors would like to thank the financial support of the Royal Society International Exchanges Award IES$\backslash$R1$\backslash$231122.} }
\maketitle
\begin{abstract}
This paper presents the distributed Poisson multi-Bernoulli (PMB)
filter based on the generalised covariance intersection (GCI) fusion
rule for distributed multi-object filtering. Since the exact GCI fusion
of two PMB densities is intractable, we derive a principled approximation.
Specifically, we approximate the power of a PMB density as an unnormalised
PMB density, which corresponds to an upper bound of the PMB density.
Then, the GCI fusion rule corresponds to the normalised product of
two unnormalised PMB densities. We show that the result is a Poisson
multi-Bernoulli mixture (PMBM), which can be expressed in closed form.
Future prediction and update steps in each filter preserve the PMBM
form, which can be projected back to a PMB density before the next
fusion step. Experimental results show the benefits of this approach
compared to other distributed multi-object filters.
\end{abstract}


\section{Introduction}

Multiple object filtering (MOF) aims to determine the unknown and
time-varying number of objects and their states based on noisy sensor
data. This field has many applications including autonomous vehicles
\cite{Wei24}, space surveillance \cite{Cataldo20} and marine traffic
monitoring \cite{Hem24}. MOF can be performed by a number of collaborative
agents that sense the environment and then fuse their information
to obtain a better understanding of the scene \cite{Su24}. Multi-agent
MOF has several advantages over single-agent MOF such as increased
robustness, improved performance and the possibility to monitor larger
areas \cite{Wang21c}.

MOF is often addressed using probabilistic inference via random finite
sets (RFSs) \cite{Mahler_book14}. In this approach, all information
about the current set of objects is contained in the density of the
set of objects given all measurements up to the current time step
(posterior density). A number of single-agent MOF algorithms that
compute or approximate the posterior density under different modelling
assumptions have been proposed. For instance, MOF algorithms based
on a labelled RFS (which require a multi-Bernoulli birth model with
uniquely labelled objects upon birth) are the generalised labelled
multi-Bernoulli (GLMB) filter \cite{Vo17} and the labelled multi-Bernoulli
(LMB) filter \cite{Reuter14}. MOF algorithms based on a Poisson point
process (PPP) birth model without object labelling are the probability
hypothesis density filter \cite{Mahler_book14}, the Poisson multi-Bernoulli
mixture (PMBM) filter \cite{Williams15b,Angel18_b} and Poisson multi-Bernoulli
(PMB) filters \cite{Williams15b,Williams15}. PMBM and PMB filters
can also be adapted to multi-Bernoulli birth model \cite{Angel19_e}. 

To perform multi-agent MOF, there are two main rules to fuse the information
contained in two multi-object densities (without known dependencies)
in a principled manner: generalised covariance intersection (GCI)
\cite{Mahler00} and arithmetic average (AA) \cite{Li20}. Importantly,
both GCI and AA are optimal in the sense of minimising a weighted
sum of Kullback-Leibler divergences (KLDs) w.r.t. the fusing densities
\cite{Li20,Gao20c}.  For MOF algorithms with labelled objects, the
GCI rule has been used for general multi-object densities, GLMB and
LMB densities in \cite{Li19c} and also for LMB densities in \cite{Kropfreiter20b,Shen22b}.
The AA fusion rule has been applied to GLMB and LMB densities in \cite{Gao20}.

A theoretical challenge with distributed fusion with labelled objects
is that the direct application of the GCI and AA fusion rules can
be problematic as different agents may have assigned different labels
to the same potential object.  Distributed labelled RFS algorithms
therefore require external procedures to resolve label mismatches
before applying the GCI of AA fusion rules. These procedures, such
as explicit label-to-label association or relabelling steps \cite{Li19c}\cite{Gao20},
 are not part of the original GCI/AA formulation, but must be performed
beforehand to ensure label consistency across agents. Another strategy
to sort out label mismatches in practice is to first remove the labels,
perform the fusion and then add them afterwards, for example, as done
for GLMB/LMB densities in \cite{Li18} and in the hybrid LMB and probability
hypothesis density (PHD) fusion algorithm in \cite{Wei24b}. The label
mismatch problem can also complicate the standard use of labels in
labelled RFS theory for sequential trajectory estimation, though it
is always possible to recover trajectory information from the sequence
of filtered densities on the set of unlabelled objects \cite{Xia22b}.

Fusion of multi-object densities without object labelling is not affected
by the theoretical limitation of label mismatches. With the AA rule,
distributed AA-PHD filters have been proposed in \cite{Li19b,Gao20c}
and a distributed AA-PMBM filter in \cite{Li23b}. With the GCI rule,
distributed GCI-PHD filters were proposed in \cite{Uney13,Battistelli13,Kim20,Wang22b},
a distributed GCI Bernoulli filter was proposed in \cite{Guldogan14},
and a distributed multi-Bernoulli filter was proposed in \cite{Wang17}.
A fusion rule based on GCI is applied to a PMBM filter for extended
objects in \cite{Lv24}, but it does not actually perform GCI on the
full PMBM as it only fuses single-object density parameters in an
ad-hoc manner. In \cite{Frohle20}, a fusion rule for PMBs based on
minimising an upper bound of the GCI-KLD cost function \cite{Li20}
is proposed. This fusion rule requires that the number of Bernoulli
and PPP components is the same for all agents, which is achieved by
dividing the PPP into different PPPs. The fusion rule has a degree
of freedom that is the choice of the permutations that assign Bernoulli
and PPP components across all agents. These permutations are chosen
by minimising a symmetric KLD between the single-object densities
of the Bernoulli/PPP components. 

\begin{figure}
\begin{centering}
\includegraphics{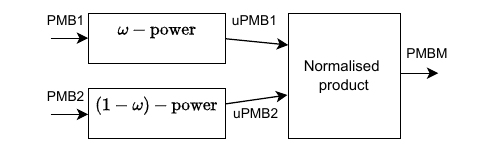}
\par\end{centering}
\caption{\label{fig:Approximate-GCI-fusion}Approximate GCI fusion rule (with
parameter $\omega\in\left[0,1\right]$) for two PMB densities. The
$\omega$ and $1-\omega$ powers of the PMB densities (PMB1, PMB2)
are upper bounded by unnormalised PMB densities (uPMB1, uPMB2). The
normalised product of these is a PMBM. The PMBM can then be projected
back to a PMB \cite{Williams15b,Williams15}.}
\end{figure}

The main contribution of this paper is the derivation of a principled
mathematical approximation of the GCI fusion rule for two PMB densities,
and the resulting distributed MOF algorithms. The only approximation
involved is in the computation of the power of a PMB, which we approximate
by an unnormalised PMB density with a closed-form expression. This
approximation results from approximating the power of a sum by a sum
of powers, which is an upper bound and a standard approximation in
GCI fusion \cite{Battistelli13}. Then, the GCI rule consists of the
calculation of the normalised product of these unnormalised PMB densities.
We show in this paper that this product is a PMBM density with a closed-form
expression, see Figure \ref{fig:Approximate-GCI-fusion} for a diagram.
This procedure naturally assigns potential objects in the two densities,
with the possibility of leaving them unassigned, without the need
of external assignment procedures. We also show that the proposed
GCI-PMB fusion rule minimises a lower bound on the GCI weighted sum
of KLDs. A second contribution  is the Gaussian implementation of
the GCI-PMB fusion rule. A third contribution is the extension of
the proposed fusion rule to sensors with limited fields of view (FoVs).
Simulation results show the benefits of the proposed fusion rule.

The rest of the paper is organised as follows. Section \ref{sec:Problem-formulation}
presents the problem formulation. The approximate GCI fusion rule
for two PMBs is derived in Section \ref{sec:Approximate-GCI-Fusion}.
Section \ref{sec:Gaussian-GCI-PMB-implementation} addresses the Gaussian
implementation of the GCI-PMB fusion rule. Section \ref{sec:Application-limited_Fov}
explains how to extend this fusion rule to sensors with limited FoVs.
Lastly, simulation results and conclusions are provided in Sections
\ref{sec:Simulations} and \ref{sec:Conclusions}, respectively.

\section{Problem formulation\label{sec:Problem-formulation}}

This paper deals with distributed multi-object filtering using PMB
filters. In particular, we consider that there are a number of sensors/nodes/agents
in the scenario, each independently performing its own PMB filtering.
 In this context, the aim of this paper is to develop the GCI fusion
for two PMB densities. The result can be applied to the information
fusion for any number of nodes via sequential fusion \cite{Gunay16}.

Let $x\in\mathbb{R}^{n_{x}}$ be a single-object state and $X$ be
the set of object states, where $X\in\mathcal{F}\left(\mathbb{R}^{n_{x}}\right)$,
which denotes the set of all finite subsets of $\mathbb{R}^{n_{x}}$.
We consider two nodes, indexed by $s\in\{1,2\}$. After processing
their measurements, node $s$ represents the information on $X$ as
a PMB density \cite{Williams15b}
\begin{align}
f_{s}\left(X\right) & =\sum_{X^{0}\uplus...\uplus X^{n_{s}}=X}f_{s}^{\mathrm{p}}\left(X^{0}\right)\prod_{i=1}^{n_{s}}f_{s}^{i}\left(X^{i}\right),\label{eq:PMB_density}
\end{align}
where the sum goes through all mutually disjoint sets $X^{0},...,X^{n_{s}}$
such that their union is $X$, $n_{s}$ is the number of Bernoulli
components, $f_{s}^{\mathrm{p}}\left(\cdot\right)$ is the PPP density
and $f_{s}^{i}\left(\cdot\right)$ is the $i$-th Bernoulli density.
These densities have the form
\begin{align}
f_{s}^{\mathrm{p}}\left(X\right) & =e^{-\int\lambda_{s}\left(x\right)dx}\left[\lambda_{s}\right]^{X}\\
f_{s}^{i}\left(X\right) & =\begin{cases}
1-r_{s}^{i} & X=\emptyset\\
r_{s}^{i}p_{s}^{i}\left(x\right) & X=\left\{ x\right\} \\
0 & \mathrm{otherwise},
\end{cases}
\end{align}
where $\lambda_{s}\left(\cdot\right)$ is the PPP intensity, $\left[\lambda_{s}\right]^{X}=\prod_{x\in X}\lambda_{s}\left(x\right)$,
with $\left[\lambda_{s}\right]^{\emptyset}=1$, and $r_{s}^{i}$ and
$p_{s}^{i}\left(\cdot\right)$ are the probability of existence and
single-object density of the $i$-th Bernoulli component. Thus, a
PMB density can be compactly described by the PPP intensity $\lambda_{s}\left(\cdot\right)$
and the parameters of its $n_{s}$ Bernoulli components, i.e., the
existence probabilites $r_{s}^{i}$ and the associated single-object
densities $p_{s}^{i}\left(\cdot\right)$.

The GCI fusion rule yields a density \cite{Mahler00}
\begin{align}
f\left(X\right) & =\frac{\left(f_{1}\left(X\right)\right)^{\omega}\left(f_{2}\left(X\right)\right)^{1-\omega}}{\int\left(f_{1}\left(X\right)\right)^{\omega}\left(f_{2}\left(X\right)\right)^{1-\omega}\delta X}\label{eq:GCI_fusion_rule}
\end{align}
where the denominator is the set integral of the numerator and ensures
that $f\left(\cdot\right)$ integrates to one. The computation of
the GCI rule for PMB densities is intractable. The aim of this paper
is to provide an approximation to this fusion rule.

\section{Approximate GCI Fusion of two PMBs\label{sec:Approximate-GCI-Fusion}}

This section presents the proposed approximation to the GCI fusion
rule of two PMBs, see (\ref{eq:GCI_fusion_rule}). Section \ref{subsec:PMB-approximation-power}
presents the unnormalised PMB approximation of the power of a PMB.
Section \ref{subsec:Normalised-product-PMBs} presents the main results
of this paper showing that the normalised product of two PMB densities
results in a PMBM density, which is then the result of the proposed
approximation to GCI fusion rule. Section \ref{subsec:PMB-projection}
reviews several approaches to project a PMBM density to a PMB density,
which is required for subsequent fusion steps. Section \ref{subsec:Minimisation-weighted-sum-KLDs}
proves that the proposed fusion rule minimises a lower bound on the
GCI weighted sum of KLDs. Finally, we discuss important aspects of
the fusion rule in Section \ref{subsec:Discussion}.

\subsection{PMB approximation of the power of a PMB\label{subsec:PMB-approximation-power}}

The computation of the power of a PMB is intractable. Nevertheless,
there is the following upper bound to the power of a PMB \cite[Eq. (2.12.2)]{Hardy_book34}
that holds for $\omega\in\left[0,1\right]$
\begin{align}
\left(f_{1}\left(X\right)\right)^{\omega} & =\left(\sum_{X^{0}\uplus...\uplus X^{n_{1}}=X}f_{1}^{\mathrm{p}}\left(X^{0}\right)\prod_{i=1}^{n_{1}}f_{1}^{i}\left(X^{i}\right)\right)^{\omega}\\
 & \leq\sum_{X^{0}\uplus...\uplus X^{n_{1}}=X}\left(f_{1}^{\mathrm{p}}\left(X^{0}\right)\right)^{\omega}\prod_{i=1}^{n_{1}}\left(f_{1}^{i}\left(X^{i}\right)\right)^{\omega}.\label{eq:upper_bound}
\end{align}
That is, the power of a PMB density is upper bounded by an unnormalised
PMB density, which can then be used as an approximation to the PMB.
This type of approximation is standard in GCI fusion, for instance,
it was used for Gaussian mixtures in \cite[Eq. (41)]{Battistelli13}
and for multi-Bernoulli densities in \cite[Prop. 2]{Wang17}. In particular,
the approximation is accurate if the regions of the single-object
space where the main mass of the densities $f_{1}^{\mathrm{p}}\left(\cdot\right)$,
$f_{1}^{1}\left(\cdot\right)$,..., $f_{1}^{n_{1}}\left(\cdot\right)$
lie are non-overlapping \cite{Wang17}. In turn, this is more likely
to happen in situations with separated objects, high probability of
detection and low measurement noise. The resulting form of the power
of a PMB is provided in the following lemma.
\begin{lem}
\label{lem:Omega_power_PMB}Let $\omega\in\left[0,1\right]$, the
$\omega$-power of a PMB density $f_{1}\left(\cdot\right)$ of the
form (\ref{eq:PMB_density}) is approximated by the upper bound (\ref{eq:upper_bound})
resulting in 
\begin{align}
\left(f_{1}\left(X\right)\right)^{\omega} & \approx\sum_{X^{0}\uplus...\uplus X^{n_{1}}=X}\left(f_{1}^{\mathrm{p}}\left(X^{0}\right)\right)^{\omega}\prod_{i=1}^{n_{1}}\left(f_{1}^{i}\left(X^{i}\right)\right)^{\omega}\label{eq:approximation_power_PMB}\\
 & =\alpha_{1}q_{1}\left(X\right)\label{eq:approximation_power_PMB2}
\end{align}
where $\alpha$ is a proportionality constant
\begin{align}
\alpha_{1} & =\frac{e^{-\omega\int\lambda_{1}\left(x\right)dx}}{e^{-\int\lambda_{1}^{\omega}\left(x\right)dx}}\nonumber \\
 & \quad\times\prod_{i=1}^{n_{1}}\left[\left(1-r_{1}^{i}\right)^{\omega}+\left(r_{1}^{i}\right)^{\omega}\int\left(p_{1}^{i}\left(x\right)\right)^{\omega}dx\right]\label{eq:alpha_1}
\end{align}
 and $q_{1}\left(\cdot\right)$ is a PMB density
\begin{align}
q_{1}\left(X\right) & =\sum_{X^{0}\uplus...\uplus X^{n_{1}}=X}q_{1}^{\mathrm{p}}\left(X^{0}\right)\prod_{i=1}^{n_{1}}q_{1}^{i}\left(X\right)\label{eq:q_1}
\end{align}
where the PPP density $q_{1}^{\mathrm{p}}\left(\cdot\right)$ has
intensity
\begin{align}
\lambda_{q,1}\left(x\right) & =\left(\lambda_{1}\left(x\right)\right)^{\omega}\label{eq:intensity_q_sensor1}
\end{align}
and the $i$-th Bernoulli density $q_{1}^{i}\left(\cdot\right)$ has
probability of existence and single-object density given by
\begin{align}
r_{q,1}^{i} & =\frac{\left(r_{1}^{i}\right)^{\omega}\int\left(p_{1}^{i}\left(x\right)\right)^{\omega}dx}{\left(1-r_{1}^{i}\right)^{\omega}+\left(r_{1}^{i}\right)^{\omega}\int\left(p_{1}^{i}\left(x\right)\right)^{\omega}dx}\label{eq:prob_exist_q_sensor1}\\
p_{q,1}^{i}\left(x\right) & =\frac{\left(p_{1}^{i}\left(x\right)\right)^{\omega}}{\int\left(p_{1}^{i}\left(x\right)\right)^{\omega}dx}.\label{eq:single_target_q_sensor1}
\end{align}
\end{lem}
The proof of Lemma \ref{lem:Omega_power_PMB} is provided in Appendix
\ref{sec:Proof_power_PMB}.  Performing an analogous approximation
with $\left(f_{2}\left(X\right)\right)^{1-\omega}$, the fused density
(\ref{eq:GCI_fusion_rule}) is given by the normalised product of
two PMBs.

\subsection{Normalised product of two PMB densities\label{subsec:Normalised-product-PMBs}}

Plugging the approximation in Lemma \ref{lem:Omega_power_PMB} for
$\left(f_{1}\left(\cdot\right)\right)^{\omega}$ and $\left(f_{2}\left(\cdot\right)\right)^{1-\omega}$,
which yield the PMBs $q_{1}\left(\cdot\right)$ and $q_{2}\left(\cdot\right)$,
into (\ref{eq:GCI_fusion_rule}), the fused density becomes the normalised
product of two PMBs
\begin{align}
q\left(X\right) & =\frac{q_{1}\left(X\right)q_{2}\left(X\right)}{\int q_{1}\left(X\right)q_{2}\left(X\right)\delta X}\nonumber \\
 & \propto\sum_{X^{0}\uplus...\uplus X^{n_{1}}=X}q_{1}^{\mathrm{p}}\left(X^{0}\right)\prod_{i=1}^{n_{1}}q_{1}^{i}\left(X^{i}\right)\nonumber \\
 & \times\sum_{Y^{0}\uplus...\uplus Y^{n_{2}}=X}q_{2}^{\mathrm{p}}\left(Y^{0}\right)\prod_{i=1}^{n_{2}}q_{2}^{i}\left(Y^{i}\right).\label{eq:normalised_product_PMBs}
\end{align}

In this section, we provide a closed-form expression for this product.
As we will see, the normalised product of two PMB densities results
in a PMBM, where the mixture considers all possible Bernoulli-to-Bernoulli
association hypotheses, with the possibility of leaving Bernoulli
components unassigned. The Bernoulli components that remain unassigned
are associated with the PPP of the other density. 

\subsubsection{Fused PMBM via assignment sets}

Before providing the expression of the resulting PMBM, we first explain
the notation. Given two real-valued functions $f\left(\cdot\right)$
and $g\left(\cdot\right)$, we use the inner product notation
\begin{align}
\left\langle f,g\right\rangle  & =\int f\left(x\right)g\left(x\right)dx.
\end{align}
We represent the Bernoulli-to-Bernoulli associations using an assignment
set $\gamma$ between $\left\{ 1,...,n_{1}\right\} $ and $\left\{ 1,...,n_{2}\right\} $.
That is, $\gamma\subseteq\left\{ 1,...,n_{1}\right\} \times\left\{ 1,...,n_{2}\right\} $
is such that if $\left(i,j\right),\left(i,j'\right)\in\gamma$ then
$j=j'$, and if $\left(i,j\right),\left(i',j\right)\in\gamma$ then
$i=i'$ \cite{Rahmathullah17}. These properties ensure that every
$i$ and $j$ gets at most one assignment. The set of all possible
$\gamma$ is $\Gamma$. It should be noted that the number of possible
assignments is
\begin{align}
|\Gamma| & =\sum_{p=0}^{\min\left(n_{1},n_{2}\right)}p!\left(\begin{array}{c}
n_{1}\\
p
\end{array}\right)\left(\begin{array}{c}
n_{2}\\
p
\end{array}\right)\label{eq:number_assignments}
\end{align}
where $p$ goes through the number of assigned objects.

In addition, for a given $\gamma$ , let $\gamma_{1}^{u,s},...,\gamma_{n_{s}-|\gamma|}^{u,s}$
denote the indices of the unassigned Bernoulli components in density
$s\in\{1,2\}$. We also write the assignment set as $\gamma=\left\{ \gamma_{1},...,\gamma_{|\gamma|}\right\} $,
$\gamma_{i}=\left(\gamma_{i}(1),\gamma_{i}(2)\right)$. The normalised
product of two PMB densities is then given in the following Theorem. 
\begin{thm}
\label{thm:Product-PMBs}The normalised product of the PMB densities
$q_{1}\left(\cdot\right)$ and $q_{2}\left(\cdot\right)$ in (\ref{eq:normalised_product_PMBs})
is a PMBM of the form
\begin{align}
q\left(X\right) & =\sum_{U\uplus W=X}q^{\mathrm{p}}\left(U\right)q^{\mathrm{mbm}}\left(W\right)\label{eq:PMBM_result}
\end{align}
where the PPP density is
\begin{align}
q^{\mathrm{p}}\left(U\right) & =e^{-\int\lambda_{q,1}(x)\lambda_{q,2}(x)dx}\left[\lambda_{q,1}\cdot\lambda_{q,2}\right]^{U}.\label{eq:PPP_result}
\end{align}
The MBM density is
\begin{align}
q^{\mathrm{mbm}}\left(W\right) & =\sum_{\gamma\in\Gamma}\rho_{\gamma}\sum_{X^{1}\uplus...\uplus X^{n_{1}}\uplus Y^{\gamma_{1}^{u,2}}\uplus...\uplus Y^{\gamma_{n_{2}-|\gamma|}^{u,2}}=W}\nonumber \\
 & \times\prod_{i=1}^{|\gamma|}q_{\gamma_{i}(1),\gamma_{i}(2)}\left(X^{\gamma_{1}(1)}\right)\nonumber \\
 & \times\prod_{i=1}^{n_{1}-|\gamma|}q_{\gamma_{i}^{u,1},0}\left(X^{\gamma_{i}^{u,1}}\right)\prod_{i=1}^{n_{2}-|\gamma|}q_{0,\gamma_{i}^{u,2}}\left(Y^{\gamma_{i}^{u,2}}\right)\label{eq:MBM_result}
\end{align}
where the weight $\rho_{\gamma}$ of assignment $\gamma$ is the product
of the weights of each Bernoulli component
\begin{align}
\rho_{\gamma} & \propto\left[\prod_{i=1}^{|\gamma|}\rho_{\gamma_{i}(1),\gamma_{i}(2)}\right]\left[\prod_{i=1}^{n_{1}-|\gamma|}\rho_{\gamma_{i}^{u,1},0}\right]\left[\prod_{i=1}^{n_{2}-|\gamma|}\rho_{0,\gamma_{i}^{u,2}}\right]
\end{align}
and $q_{i,j}\left(\cdot\right)$ is a Bernoulli density with probability
of existence $r_{i,j}$ single-object density $p_{i,j}\left(\cdot\right)$
and weight $\rho_{i,j}$, which are calculated as follows. For assigned
Bernoulli densities, $i\in\left\{ 1,...,n_{1}\right\} $ and $j\in\left\{ 1,...,n_{2}\right\} $,
we have that 
\begin{align}
r_{i,j} & =\frac{r_{q,1}^{i}r_{q,2}^{j}\left\langle p_{q,1}^{i},p_{q,2}^{j}\right\rangle }{\rho_{i,j}}\\
p_{i,j}\left(x\right) & =\frac{p_{q,1}^{i}\left(x\right)p_{q,2}^{j}\left(x\right)}{\left\langle p_{q,1}^{i},p_{q,2}^{j}\right\rangle }\\
\rho_{i,j} & =\left(1-r_{q,1}^{i}\right)\left(1-r_{q,2}^{j}\right)+r_{q,1}^{i}r_{q,2}^{j}\left\langle p_{q,1}^{i},p_{q,2}^{j}\right\rangle .
\end{align}
For unassigned Bernoulli densities in the first PMB density, with
$i\in\left\{ 1,...,n_{1}\right\} $, we have
\begin{align}
r_{i,0} & =\frac{r_{q,1}^{i}\left\langle p_{q,1}^{i},\lambda_{q,2}\right\rangle }{\rho_{i,0}}\\
p_{i,0}\left(x\right) & =\frac{p_{q,1}^{i}\left(x\right)\lambda_{q,2}\left(x\right)}{\left\langle p_{q,1}^{i},\lambda_{q,2}\right\rangle }\\
\rho_{i,0} & =1-r_{q,1}^{i}+r_{q,1}^{i}\left\langle p_{q,1}^{i},\lambda_{q,2}\right\rangle .
\end{align}
For unassigned Bernoulli densities in the second PMB density, with
$j\in\left\{ 1,...,n_{2}\right\} $, we have
\begin{align}
r_{0,j} & =\frac{r_{q,2}^{j}\left\langle p_{q,2}^{j},\lambda_{q,1}\right\rangle }{\rho_{0,j}}\\
p_{0,j}\left(x\right) & =\frac{p_{q,2}^{j}\left(x\right)\lambda_{q,1}\left(x\right)}{\left\langle p_{q,2}^{j},\lambda_{q,1}\right\rangle }\\
\rho_{0,j} & =1-r_{q,2}^{j}+r_{q,2}^{j}\left\langle p_{q,2}^{j},\lambda_{q,1}\right\rangle .
\end{align}
\end{thm}

\begin{figure}
\begin{centering}
\includegraphics[scale=0.6]{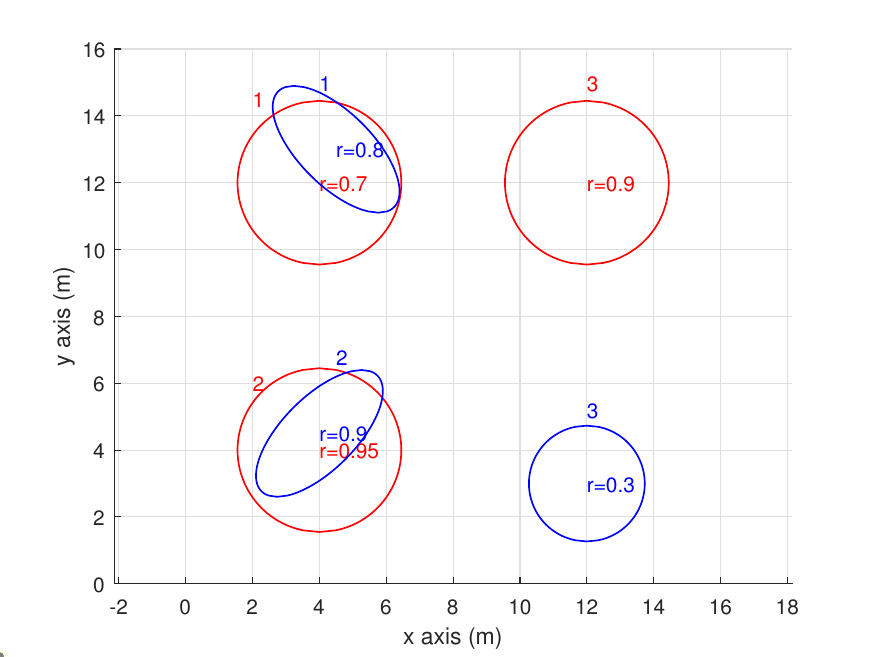}
\par\end{centering}
\begin{centering}
\includegraphics[scale=0.6]{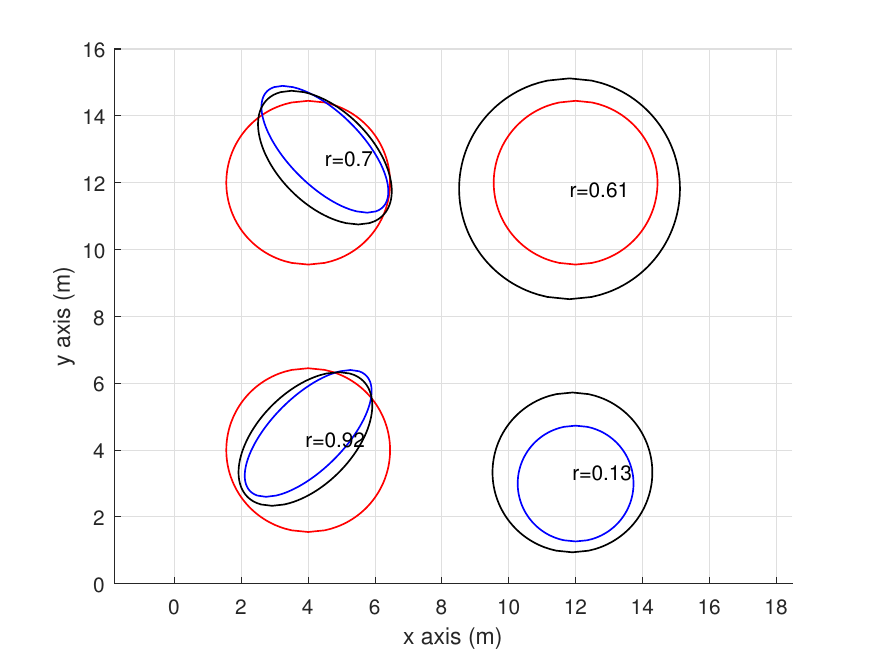}
\par\end{centering}
\caption{\label{fig:Example-fusion}Example fusion of two PMBs $q_{1}\left(\cdot\right)$
(red) and $q_{2}\left(\cdot\right)$ (blue). Both PMBs have three
Bernoulli components, indexed by 1, 2 and 3 (top figure). The probability
of existence is shown at the center of the 3-$\sigma$ ellipse representing
the mean and covariance matrix. The bottom figure shows the PMB (black)
in the most likely hypothesis of the fused PMBM. }
\end{figure}

We prove Theorem \ref{thm:Product-PMBs} in Appendix \ref{sec:Proof_Theorem1}. 
\begin{example}
Let us consider the PMBs $q_{1}\left(\cdot\right)$ and $q_{2}\left(\cdot\right)$
shown in Figure \ref{fig:Example-fusion} (top). The single-object
states contain 2-D positions.  Both PMBs have three Bernoulli components,
with Gaussian single-object densities represented by confidence ellipses.
The PPP intensity in both cases is Gaussian with weight 1, mean $[10,10]^{T}$
and covariance matrix $20I_{2}$, covering the whole area of interest.
Using (\ref{eq:number_assignments}), the number of possible assignments
is $|\Gamma|=34$. The global hypothesis (assignment set $\gamma$)
with highest weight is $\gamma=\left\{ \left(1,1\right),(2,2)\right\} $.
In this hypothesis, the third Bernoulli component in each PMB is left
unassigned (associated to the PPP), resulting in $\gamma_{1}^{u,1}=3$
and $\gamma_{1}^{u,2}=3$. The fused PMB of this global hypothesis
has four Bernoulli components and is shown in Figure \ref{fig:Example-fusion}
(bottom). The assigned Bernoulli components give rise to fused Bernoulli
components merging the information from the original Bernoulli components.
The unassigned Bernoulli components are fused with the PPP adjusting
their probabilities of existence, means and covariance matrices.

\end{example}

\subsubsection{Fused PMBM in track-oriented form}

Theorem \ref{thm:Product-PMBs} can also be expressed in a track-oriented
form using standard PMBM filtering notation \cite{Williams15b}. This
helps the implementation of the filter as it is possible to leverage
available PMBM filter implementations that make use of the track-oriented
form. To do this, we think of $q_{1}\left(\cdot\right)$ as the prior
PMB and the Bernoulli components of $q_{2}\left(\cdot\right)$ as
the received set of measurements. The fused PMBM then has $n_{1}+n_{2}$
Bernoulli components, obtained by appending the Bernoulli components
of $q_{2}\left(\cdot\right)$ to those of $q_{1}\left(\cdot\right)$.
In this notation, the assignment set $\gamma$ is represented with
a global hypothesis $a=\left(a^{1},...,a^{n_{1}+n_{2}}\right)$, where
$a^{i}\in\left\{ 1,...,h^{i}\right\} $ is the index to the local
hypothesis for the $i$-th Bernoulli and $h^{i}$ is the number of
local hypotheses. A local hypothesis assigns the $i$-th Bernoulli
to another Bernoulli in $q_{2}\left(\cdot\right)$ or leaves it unassigned.
With this notation, the MBM component of the PMBM can be written as
\begin{align}
q^{\mathrm{mbm}}\left(W\right) & \propto\sum_{a\in\mathcal{A}_{k}}\sum_{X^{1}\uplus...\uplus X^{n_{1}+n_{2}}=W}\prod_{i=1}^{n_{1}+n_{2}}\rho^{i,a^{i}}q^{i,a^{i}}\left(X^{i}\right)\label{eq:MBM_TO_form}
\end{align}
where $\mathcal{A}_{k}$ is the set of all global hypotheses, and
$\rho^{i,a^{i}}$ and $q^{i,a^{i}}\left(\cdot\right)$ are the weight
and Bernoulli density corresponding to local hypothesis $a^{i}$ in
the $i$-th Bernoulli. The density $q^{i,a^{i}}\left(\cdot\right)$
is characterised by its probability of existence $r^{i,a^{i}}$ and
its single-object density $p^{i,a^{i}}\left(\cdot\right)$.

For each local hypothesis $\left(i,a^{i}\right)$, there is a set
$\mathcal{M}\left(i,a^{i}\right)$ such that $\mathcal{M}\left(i,a^{i}\right)=\left\{ j\right\} $
if it has been assigned to the $j$-th Bernoulli in density $q_{2}\left(\cdot\right)$,
and $\mathcal{M}\left(i,a^{i}\right)=\emptyset$ if it has been unassigned.
Then, the set of all global hypotheses is \cite{Angel20_e}
\begin{align}
\mathcal{A}_{k} & =\left\{ \vphantom{\bigcup_{i=1}^{n_{k|k}}}\left(a_{1},...,a_{n_{1}+n_{2}}\right):a_{i}\in\mathbb{N}_{h_{i}},\bigcup_{i=1}^{n_{1}+n_{2}}\mathcal{M}\left(i,a_{i}\right)=\mathbb{N}_{n_{2}},\right.\nonumber \\
 & \left.\quad\mathcal{M}\left(i,a_{i}\right)\cap\mathcal{M}\left(j,a_{j}\right)=\emptyset\,\forall i\neq j\vphantom{\bigcup_{i=1}^{n_{k|k}}}\right\} \label{eq:global_hypotheses_PMB}
\end{align}
where $\mathbb{N}_{n}=\left\{ 1,...,n\right\} $. The set $\mathcal{A}_{k}$
ensures the same constraints as the assignment sets $\gamma\in\Gamma$.
The following lemma provides the track-oriented form of Theorem \ref{thm:Product-PMBs},
providing expressions for $q^{i,a^{i}}\left(\cdot\right)$, $\rho^{i,a^{i}}$
and $\mathcal{M}\left(i,a^{i}\right)$.
\begin{lem}
\label{lem:Product_PMBs_TO}The normalised product of the PMB densities
$q_{1}\left(\cdot\right)$ and $q_{2}\left(\cdot\right)$ in (\ref{eq:normalised_product_PMBs})
is a PMBM of the form (\ref{eq:PMBM_result}) where the PPP has intensity
(\ref{eq:PPP_result}). The MBM density in (\ref{eq:MBM_result})
can also be written in track-oriented form, see (\ref{eq:MBM_TO_form}),
as follows. For each Bernoulli component in PMB $q_{1}\left(\cdot\right)$,
$i\in\left\{ 1,...,n_{1}\right\} $, there are $h^{i}=n_{2}+1$ local
hypotheses, corresponding to an unassignment hypothesis and an assignment
with one of the Bernoulli components in PMB $q_{2}\left(\cdot\right)$.
The unassignment hypothesis for the $i$-th Bernoulli component is
characterised by 
\begin{align}
\mathcal{M}\left(i,1\right) & =\emptyset\\
r^{i,1} & =\frac{r_{q,1}^{i}\left\langle p_{q,1}^{i},\lambda_{q,2}\right\rangle }{\rho^{i,1}}\\
p^{i,1}\left(x\right) & =\frac{p_{q,1}^{i}\left(x\right)\lambda_{q,2}\left(x\right)}{\left\langle p_{q,1}^{i},\lambda_{q,2}\right\rangle }\\
\rho^{i,1} & =1-r_{q,1}^{i}+r_{q,1}^{i}\left\langle p_{q,1}^{i},\lambda_{q,2}\right\rangle .
\end{align}
The hypothesis assigning Bernoulli component $i\in\left\{ 1,...,n_{1}\right\} $
in $q_{1}\left(\cdot\right)$ to Bernoulli component $i\in\left\{ 1,...,n_{2}\right\} $
in $q_{2}\left(\cdot\right)$ is characterised by
\begin{align}
\mathcal{M}\left(i,1+j\right) & =\left\{ j\right\} \\
r^{1,1+j} & =\frac{r_{q,1}^{i}r_{q,2}^{j}\left\langle p_{q,1}^{i},p_{q,2}^{j}\right\rangle }{\rho^{i,1+j}}\\
p^{i,1+j}\left(x\right) & =\frac{p_{q,1}^{i}\left(x\right)p_{q,2}^{j}\left(x\right)}{\left\langle p_{q,1}^{i},p_{q,2}^{j}\right\rangle }\\
\rho^{i,1+j} & =\left(1-r_{q,1}^{i}\right)\left(1-r_{q,2}^{j}\right)\nonumber \\
 & \quad+r_{q,1}^{i}r_{q,2}^{j}\int p_{q,1}^{i}\left(x\right)p_{q,2}^{j}\left(x\right)dx.
\end{align}
Finally, for Bernoulli components created by the Bernoulli components
in $q_{2}\left(\cdot\right)$, with indices $i\in\left\{ n_{1}+1,...,n_{2}\right\} $,
there are two local hypotheses. The first local hypothesis is a non-existent
Bernoulli, parameterised by $\mathcal{M}\left(i,1\right)=\emptyset$,
$\rho^{i,1}=1$, $r^{i,1}=0$. The second local hypothesis represents
that the $j$-th Bernoulli in $q_{2}\left(\cdot\right)$ is unassigned
and is parameterised by
\begin{align}
\mathcal{M}\left(i,2\right) & =\left\{ j\right\} \\
r^{i,2} & =\frac{r_{q,2}^{j}\left\langle p_{q,2}^{j},\lambda_{q,1}\right\rangle }{\rho^{i,2}}\\
p^{i,2}\left(x\right) & =\frac{p_{q,2}^{j}\left(x\right)\lambda_{q,1}\left(x\right)}{\left\langle p_{q,2}^{j},\lambda_{q,1}\right\rangle }\\
\rho^{i,2} & =1-r_{q,2}^{j}+r_{q,2}^{j}\left\langle p_{q,2}^{j},\lambda_{q,1}\right\rangle .
\end{align}
\end{lem}
The pseudocode summarising the GCI-PMB fusion rule is provided in
Algorithm \ref{alg:GCI-PMB_pseudocode}.
\begin{center}
\begin{algorithm}
\caption{\label{alg:GCI-PMB_pseudocode}GCI-PMB fusion of two PMB densities.}

\textbf{Input: }PMB densities $f_{1}\left(\cdot\right)$ and $f_{2}\left(\cdot\right)$,
see (\ref{eq:PMB_density}), fusion parameter $\omega\in\left[0,1\right]$.

\textbf{Output: }Fused PMBM $q\left(\cdot\right)$, see (\ref{eq:PMBM_result}).

\begin{algorithmic}    

\State - Approximate $\omega$-power of $f_{1}\left(\cdot\right)$
via Lemma \ref{lem:Omega_power_PMB}, obtaining PMB $q_{1}\left(\cdot\right)$.

\State - Approximate $\left(1-\omega\right)$-power of $f_{2}\left(\cdot\right)$
via Lemma \ref{lem:Omega_power_PMB}, obtaining PMB $q_{2}\left(\cdot\right)$.

\State - Calculate the normalised product of $q_{1}\left(\cdot\right)$
and $q_{2}\left(\cdot\right)$ using Lemma \ref{lem:Product_PMBs_TO},
obtaining the fused PMBM $q\left(\cdot\right)$.

\end{algorithmic}
\end{algorithm}
\par\end{center}

\subsection{PMB projection\label{subsec:PMB-projection}}

Once this update is performed, each agent uses the fused PMBM density
to continue with the filtering recursion in PMBM form \cite{Williams15b,Angel18_b}.
When another fusion step is required, this PMBM density is projected
to a PMB density. Several projection methods can be used: 
\begin{itemize}
\item Global nearest neighbour (GNN) PMB projection, which selects the most
likely assignment \cite{Crouse16}.
\item Track-oriented PMB projection (based on Kullback-Leibler divergence
minimisation with auxiliary variables) \cite{Williams15b,Angel20_e},
implemented by selecting the $k$-best asssignments using Murty's
algorithm \cite{Murty68} or by belief propagation \cite{Williams14}. 
\item Variational PMB projection \cite{Williams15,Xia22}.
\end{itemize}

It is also possible to perform the PMB projection immediately after
fusion, following one of the methods above.

\subsection{Minimisation of the weighted sum of KLDs\label{subsec:Minimisation-weighted-sum-KLDs}}

The following lemma indicates that the GCI-PMB fusion rule minimises
a lower bound on the GCI weighted sum of KLDs.
\begin{lem}
\label{lem:Minimisation_KLD}Given the PMB densities $f_{1}\left(\cdot\right)$
and $f_{2}\left(\cdot\right)$ in (\ref{eq:PMB_density}), the approximate
GCI PMB fusion rule arising from Lemma \ref{lem:Omega_power_PMB}
and Theorem \ref{thm:Product-PMBs} provides the density $q\left(\cdot\right)$
that minimises the lower bound 
\begin{align}
L\left[q\right]= & D\left(q||\frac{q_{1}q_{2}}{\left\langle q_{1},q_{2}\right\rangle }\right)-\log\left(\alpha_{1}\alpha_{2}\left\langle q_{1},q_{2}\right\rangle \right)
\end{align}
to the GCI weighted sum of KLDs
\begin{align}
C\left[q\right] & =\omega D\left(q||f_{1}\right)+\left(1-\omega\right)D\left(q||f_{2}\right)\label{eq:weighted_KLD}
\end{align}
where $\alpha_{1}$ and $q_{1}\left(\cdot\right)$ are given by (\ref{eq:alpha_1})
and (\ref{eq:q_1}), and $\alpha_{2}$ and $q_{2}\left(\cdot\right)$
are given by same equations but using $f_{2}\left(\cdot\right)$ and
$1-\omega$ instead of $f_{1}\left(\cdot\right)$ and $\omega$.
\end{lem}
That is, the GCI fusion rule (\ref{eq:GCI_fusion_rule}) minimises
$C\left[q\right]$ whereas the proposed approximate GCI fusion rule
minimises its lower bound $L\left[q\right]\leq C\left[q\right]$.

\subsection{Discussion\label{subsec:Discussion}}

This section provides a discussion of the proposed fusion rule, covering
its benefits, its comparison with \cite{Frohle20}, and the extension
to PMBM densities. The first benefit is that the fusion rule only
requires the approximation in Lemma \ref{lem:Omega_power_PMB}, while
all the subsequent steps admit closed-form expressions. Since the
PMBM form is kept in the filtering recursion for a large number of
multi-object measurement models, including point and extended objects
\cite{Williams15b,Angel18_b,Granstrom20,Angel23}, the second benefit
is that the fused output can be directly used by each agent to continue
with the local filtering recursion. The third benefit is the interpretability
of the fused result. The fused PMBM density contains all possible
assignments between the Bernoulli components (potential objects) in
both PMBs, with the possibility of leaving them unassigned. The unassigned
Bernoulli components are fused with the PPP, which represents objects
that remain undetected, in the other filter and initiate a new track
in the fused PMBM density. This is a natural way of fusing MOF information
that intrinsically starts new tracks and performs the association
between potential objects. This is in contrast with other existing
approaches that use external methods to perform the assignment and
initiate new tracks, see for instance \cite{Li18,Li23b}. While the
PPP component of the PMB is a natural way to account for the presence
of undetected objects outside the FoV, the direct application of the
proposed GCI fusion rule to sensors with limited FoVs does not yield
satisfactory results, as it can lead to an increase of uncertainty
\cite{Wang22b}. How to apply the proposed fusion rule in this case
is explained in Section \ref{sec:Application-limited_Fov}. 

The proposed fusion rule minimises a lower bound on the GCI-KLD (\ref{eq:weighted_KLD})
and returns a PMBM. In contrast, the GCI rule in \cite{Frohle20}
minimises an upper bound and returns a PMB. To do this, the fusion
in \cite{Frohle20} first divides the PPP of each agent into a number
of independent PPPs such that the total number of PPPs plus Bernoulli
components is the same for each agent. Then, the algorithm finds the
PMB that minimises an upper bound on the weighted KLD across the agents.
This KLD upper bound can freely choose the permutations that assign
PPP or Bernoulli components across different agents. The proposed
way in \cite{Frohle20} to obtain these permutations is by fusing
the information of 2 agents at a time by minimising a symmetric KLD
between the single-object densities, which can be done solving a 2D
assignment problem. On the contrary, the proposed method only requires
the approximation in Lemma \ref{lem:Omega_power_PMB} and directly
provides the weights of the PMBM global hypotheses. The most likely
hypothesis (representing a PMB) can also be computed solving a 2D
assignment problem. 

The proposed GCI fusion method can in principle be applied to PMBM
densities by making the approximation (\ref{eq:approximation_power_PMB})
to the PMBM, and then calculating the normalised product of two PMBM
densities, which would return a PMBM density (for instance, one can
apply the PMB fusion result for each pair of PMBs in the PMBM). There
are a number of challenges though. The first one is that the resulting
PMBM would not be in a track-oriented form, which is required for
a computationally efficient implementation. It may be possible to
design methods to write the fused PMBM in track-oriented form, or
perform some type of approximation to achieve this. An additional
challenge is that the computational complexity of PMBM GCI fusion
can be considerably higher than for PMB fusion due to the possibly
high number of global hypotheses. These are topics of future research. 

Finally, if there are more than two agents, the multi-agent GCI fusion
rule can be applied sequentially by adjusting the fusion weights \cite[Table I]{Gunay16}.
In our case, we must apply a PMB projection, see Section \ref{subsec:PMB-projection},
before each fusion step.

\section{Gaussian GCI-PMB implementation\label{sec:Gaussian-GCI-PMB-implementation}}

In this section, we provide the Gaussian implementation of the fusion
rule. In the Gaussian implementation, we assume that the PPP intensity
of the $s$-th node is an unnormalised Gaussian mixture
\begin{align}
\lambda_{s}\left(x\right) & =\sum_{j=1}^{n_{s}^{p}}w_{s}^{p,j}\mathcal{N}\left(x;\overline{x}_{s}^{p,j},P_{s}^{p,j}\right),\label{eq:PPP_intensity_GM}
\end{align}
where $n_{s}^{p}$ is the number of components, $w_{s}^{p,j}>0$,
$\overline{x}_{s}^{p,j}$ and $P_{s}^{p,j}$ are the weight, mean
and covariance matrix of the $j$-th component, respectively.

The single-object density of the $i\in\{1,...,n_{s}\}$ Bernoulli
of the $s$-th node is Gaussian
\begin{align}
p_{s}^{i}\left(x\right) & =\mathcal{N}\left(x;\overline{x}_{s}^{i},P_{s}^{i}\right)\label{eq:single_target_density_Gaussian}
\end{align}
where $\overline{x}_{s}^{i}$ and $P_{s}^{i}$ are the mean and covariance
matrix, respectively. Intensities and densities of the forms (\ref{eq:PPP_intensity_GM})
and (\ref{eq:single_target_density_Gaussian}) naturally arise in
PMBM/PMB filtering with linear Gaussian dynamic and measurement models
under constant probabilities of survival and detection \cite{Williams15b,Angel18_b}.

Section \ref{subsec:PMB-approximation-Gaussian} explains the PMB
approximation of the power of a PMB, Section \ref{subsec:GCI-rule_Gaussian}
explains the GCI rule, and Section \ref{subsec:Practical-aspects}
discusses some practical aspects.

\subsection{PMB approximation of the power of a PMB\label{subsec:PMB-approximation-Gaussian}}

\subsubsection{Power of PPP intensity}

Substituting (\ref{eq:PPP_intensity_GM}) into (\ref{eq:intensity_q_sensor1}),
the PPP intensity for node 1 is
\begin{align}
\lambda_{q,1}\left(x\right) & =\left(\sum_{j=1}^{n_{1}^{p}}w_{1}^{p,j}\mathcal{N}\left(x;\overline{x}_{1}^{p,j},P_{1}^{p,j}\right)\right)^{\omega}\nonumber \\
 & \approx\sum_{j=1}^{n_{1}^{p}}\left(w_{1}^{p,j}\right)^{\omega}\left(\mathcal{N}\left(x;\overline{x}_{1}^{p,j},P_{1}^{p,j}\right)\right)^{\omega}\nonumber \\
 & =\sum_{j=1}^{n_{1}^{p}}\left(w_{1}^{p,j}\right)^{\omega}\kappa\left(\omega,P_{1}^{p,j}\right)\mathcal{N}\left(x;\overline{x}_{1}^{p,j},\frac{P_{1}^{p,j}}{\omega}\right)\label{eq:Gaussian_mixture_intensity1}
\end{align}
where
\begin{align}
\kappa\left(\omega,P\right) & =\frac{|2\pi P\omega^{-1}|^{1/2}}{|2\pi P|^{\omega/2}}
\end{align}
and we have applied the standard approximation for the exponential
of a mixture in \cite[Eq. (41)]{Battistelli13}. Similarly, for the
intensity of node 2, we have that
\begin{align}
\lambda_{q,2}\left(x\right) & \approx\sum_{j=1}^{n_{2}^{p}}\left(w_{2}^{p,j}\right)^{1-\omega}\kappa\left(1-\omega,P_{2}^{p,j}\right)\nonumber \\
 & \times\mathcal{N}\left(x;\overline{x}_{2}^{p,j},\frac{P_{2}^{p,j}}{1-\omega}\right).\label{eq:Gaussian_mixture_intensity2}
\end{align}

\subsubsection{Power of Bernoulli density}

Substituting (\ref{eq:single_target_density_Gaussian}) into (\ref{eq:prob_exist_q_sensor1})
and (\ref{eq:single_target_q_sensor1}) yields
\begin{align}
r_{q,1}^{i} & =\frac{\left(r_{1}^{i}\right)^{\omega}\kappa\left(\omega,P_{1}^{i}\right)}{\left(1-r_{1}^{i}\right)^{\omega}+\left(r_{1}^{i}\right)^{\omega}\kappa\left(\omega,P_{1}^{i}\right)}\label{eq:prob_exist_q_sensor1_Gaussian}\\
p_{q,1}^{i}\left(x\right) & =\mathcal{N}\left(x;\overline{x}_{s}^{i},\frac{P_{s}^{i}}{\omega}\right)\label{eq:single_target_q_sensor1_Gaussian}
\end{align}
where we have used the fact that \cite[Eq. (36)]{Battistelli13}
\begin{align}
\int\left(p_{1}^{i}\left(x\right)\right)^{\omega}dx & =\kappa\left(\omega,P_{1}^{i}\right).
\end{align}

\subsection{GCI rule\label{subsec:GCI-rule_Gaussian}}

The GCI fusion rule for two Gaussian PMB densities is obtained as
indicated in the next lemma. 
\begin{lem}
Let us consider two PMB densities, denoted by $q_{s}\left(\cdot\right)$
with index $s\in\{1,2\}$, with Gaussian mixture PPP intensity
\begin{align}
\lambda_{q,s}\left(x\right) & =\sum_{j=1}^{n_{q,s}^{p}}w_{q,s}^{p,j}\mathcal{N}\left(x;\overline{x}_{q,s}^{p,j},P_{q,s}^{p,j}\right)\label{eq:lambda_q_s_Gaussian}
\end{align}
and MB characterised by probabilities of existence $r_{q,s}^{i}$
and single-object densities
\begin{align}
p_{q,s}^{i}\left(x\right) & =\mathcal{N}\left(x;\overline{x}_{q,s}^{i},P_{q,s}^{i}\right)\label{eq:p_q_s_Gaussian}
\end{align}
for $i\in\{1,...,n_{s}\}$. The normalised product (\ref{eq:normalised_product_PMBs})
of these PMB densities is the PMBM in Theorem \ref{thm:Product-PMBs}
with density (\ref{eq:PMBM_result}). The fused PPP intensity is then
given by
\begin{align}
 & \lambda_{q,1}\left(x\right)\cdot\lambda_{q,2}\left(x\right)\nonumber \\
 & \quad=\sum_{j_{1}=1}^{n_{q,1}^{p}}\sum_{j_{2}=1}^{n_{q,2}^{p}}w_{q,1}^{p,j_{1}}w_{q,2}^{p,j_{2}}\alpha_{0,0}^{j_{1},j_{2}}\mathcal{N}\left(x;\overline{x}_{0,0}^{j_{1},j_{2}},P_{0,0}^{j_{1},j_{2}}\right)\label{eq:PPP_fusion_Gaussian}
\end{align}
where
\begin{align}
\alpha_{0,0}^{j_{1},j_{2}} & =\mathcal{N}\left(\overline{x}_{q,1}^{p,j_{1}};\overline{x}_{q,2}^{p,j_{2}},P_{q,1}^{p,j_{1}}+P_{q,2}^{p,j_{2}}\right)\\
P_{0,0}^{j_{1},j_{2}} & =\left(\left(P_{q,1}^{p,j_{1}}\right)^{-1}+\left(P_{q,2}^{p,j_{2}}\right)^{-1}\right)^{-1}\\
\overline{x}_{0,0}^{j_{1},j_{2}} & =P_{0,0}^{j_{1},j_{2}}\left(\left(P_{q,1}^{p,j_{1}}\right)^{-1}\overline{x}_{q,1}^{p,j_{1}}+\left(P_{q,2}^{p,j_{2}}\right)^{-1}\overline{x}_{q,2}^{p,j_{2}}\right).
\end{align}

For the assigned Bernoulli components, with $i\in\left\{ 1,...,n_{1}\right\} $
and $j\in\left\{ 1,...,n_{2}\right\} $, the resulting Bernoulli density
is parameterised by
\begin{align}
r_{i,j} & =\frac{r_{q,1}^{i}r_{q,2}^{j}\alpha^{i,j}}{\rho_{i,j}}\\
p_{i,j}\left(x\right) & =\mathcal{N}\left(x;\overline{x}^{i,j},P^{i,j}\right)\\
\rho_{i,j} & =\left(1-r_{q,1}^{i}\right)\left(1-r_{q,2}^{j}\right)+r_{q,1}^{i}r_{q,2}^{j}\alpha^{i,j}
\end{align}
where
\begin{align}
\alpha^{i,j} & =\mathcal{N}\left(\overline{x}_{q,1}^{i};\overline{x}_{q,2}^{j},P_{q,1}^{i}+P_{q,2}^{j}\right)\label{eq:alpha_ij_weight}\\
P^{i,j} & =\left(\left(P_{q,1}^{i}\right)^{-1}+\left(P_{q,2}^{j}\right)^{-1}\right)^{-1}\\
\overline{x}^{i,j} & =P^{i,j}\left(\left(P_{q,1}^{i}\right)^{-1}\overline{x}_{q,1}^{i}+\left(P_{q,2}^{j}\right)^{-1}\overline{x}_{q,2}^{j}\right).
\end{align}

For unassigned Bernoulli densities in the first PMB density, with
$i\in\left\{ 1,...,n_{1}\right\} $, we have
\begin{align}
r_{i,0} & =\frac{r_{q,1}^{i}\sum_{j=1}^{n_{q,2}^{p}}w_{q,2}^{p,j}\alpha_{i,0}^{j}}{\rho_{i,0}}\label{eq:r_i_0_Gaussian}\\
p_{i,0}\left(x\right) & \propto\sum_{j=1}^{n_{q,2}^{p}}w_{q,2}^{p,j}\alpha_{i,0}^{j}\mathcal{N}\left(x;\overline{x}_{i,0}^{j},P_{i,0}^{j}\right)\\
\rho_{i,0} & =1-r_{q,1}^{i}+r_{q,1}^{i}\sum_{j=1}^{n_{q,s}^{p}}w_{q,2}^{p,j}\alpha_{i,0}^{i,j}
\end{align}
where
\begin{align}
\alpha_{i,0}^{j} & =\mathcal{N}\left(\overline{x}_{q,1}^{i};\overline{x}_{q,2}^{p,j},P_{q,1}^{i}+P_{q,2}^{p,j}\right)\\
P_{i,0}^{j} & =\left(\left(P_{q,1}^{i}\right)^{-1}+\left(P_{q,2}^{p,j}\right)^{-1}\right)^{-1}\\
\overline{x}_{i,0}^{j} & =P_{i,0}^{j}\left(\left(P_{q,1}^{i}\right)^{-1}\overline{x}_{q,1}^{i}+\left(P_{q,2}^{p,j}\right)^{-1}\overline{x}_{q,2}^{p,j}\right).\label{eq:x_i_0_Gaussian}
\end{align}

For unassigned Bernoulli densities in the second PMB density, with
$j\in\left\{ 1,...,n_{2}\right\} $, the results are analogous to
(\ref{eq:r_i_0_Gaussian})-(\ref{eq:x_i_0_Gaussian}) with indices
changed accordingly to obtain $r_{0,j},p_{0,j}\left(\cdot\right),\rho_{0,j}$.

\end{lem}
The proof of this lemma is direct by using Theorem \ref{thm:Product-PMBs}
and the identity for the product of two Gaussian densities \cite[Eq. (11)]{Williams03}.
Finally, the PMB projection for the Gaussian implementation also does
moment matching to keep each single-object posterior as Gaussian \cite{Williams15b,Angel20_e}.

\subsection{Practical aspects\label{subsec:Practical-aspects}}

As the result of GCI-PMB fusion is a PMBM, it is important to limit
the number of global hypotheses, Bernoulli components and PPP components
for practical implementation. Each agent already performs these operations
when running its PMB/PMBM filter but it is also required when performing
the fusion rule. This section explains global hypothesis pruning,
gating, and pruning-merging of PPP components when performing GCI-PMB
fusion. 

A GCI-PMB fusion rule implementation requires pruning global hypotheses
resulting from Lemma \ref{lem:Product_PMBs_TO}. This can be done,
for example, via Murty's algorithm or variants \cite{Murty68,Pedersen08}
or Gibbs sampling \cite{Vo17}. These standard approaches can be used
because, as a result of PMB fusion, the weight of each global hypothesis
can be factorised as the product of the weights of the local association
hypotheses. Gating is a standard mechanism to lower the number of
data association hypotheses \cite{Challa_book11}. In standard PMBM
filtering, the gating between a measurement $z$ and a certain Bernoulli
component is performed using the square Mahalanobis distance $\left(z-\hat{z}\right)^{T}S^{-1}\left(z-\hat{z}\right)$
where $\hat{z}$ and $S$ are the predicted measurement and its covariance
matrix \cite{Challa_book11}. This is done because the local hypothesis
weight depends on this distance. 

In contrast, in GCI-PMB fusion, the weight of the local hypothesis
depends on $\alpha^{i,j}$ in (\ref{eq:alpha_ij_weight}) which, in
turn, depends on the square Mahalanobis distance \cite{Battistelli13}
\begin{align}
d^{i,j} & =\left(\overline{x}_{q,1}^{i}-\overline{x}_{q,2}^{j}\right)^{T}\left(P_{q,1}^{i}+P_{q,2}^{j}\right)^{-1}\left(\overline{x}_{q,1}^{i}-\overline{x}_{q,2}^{j}\right).\label{eq:Mahalanobis}
\end{align}
Given a gating threshold $\Gamma_{f}$, if $d^{i,j}<\Gamma_{f}$,
then Bernoulli $i$ can be assigned to Bernoulli $j$. The same logic
is used to assign PPP components to Bernoulli components (in the unassigned
local hypotheses).

To lower the number of PPP components, we use pruning and merging
after both the local update and the GCI fusion step. Due to the approximation
in (\ref{eq:Gaussian_mixture_intensity1}), it is particularly important
to merge similar PPP components, as the approximation improves if
the Gaussian densities are well separated \cite{Battistelli13}. In
the simulations in Section \ref{sec:Simulations}, each agent also
prunes global hypotheses and Bernoulli components after the local
update and the GCI fusion step. Due to the approximation (\ref{eq:approximation_power_PMB}),
recycling \cite{Williams12} and merging of Bernoulli components after
the GCI fusion step may also help improve performance.

\section{Application to sensors with limited FoVs\label{sec:Application-limited_Fov}}

In this section, we explain one approach to apply the proposed GCI-PMB
fusion rule with sensors with limited FoVs. The approach is based
on PMB partitioning followed by GCI-PMB fusion in different areas.
The partitioning and fusion algorithm is explained in Section \ref{subsec:Partitioning-and-fusion}.
Section \ref{subsec:Bernoulli-assignment} explains how to assign
Bernoulli components to each partition.

\subsection{Partitioning and fusion\label{subsec:Partitioning-and-fusion}}

When we consider the GCI fusion of two multi-object densities obtained
by sensors with limited FoVs, there can be an increase of uncertainty
in those areas which are observed by one sensor but not by the other
\cite{Wang22b}. This is due to the fact that GCI provides a geometric
average between two multi-object densities. Therefore, if one of them
has little information about an area, but the other has accurate information
about an area, the GCI rule would increase the uncertainty unless
we set $\omega$ to only consider the information of the sensor with
relevant information. This situation gets worse if there are multiple
agents with limited FoVs. One solution to this phenomenon proposed
in \cite{Wang22b} is that each sensor uses an uninformative density
outside its FoV. Then, independent fusion is used in the area which
is observed by one sensor but not by the other. The GCI rule is only
used for the area that is within the FoVs of both sensors.

Based on these insights, we propose the following approach to handle
limited FoVs. We first divide the single-object space into three areas
$A_{0},A_{1},A_{2}\subseteq\mathbb{R}^{n_{x}}$, where
\begin{itemize}
\item $A_{1}$ is the area that contains the FoV of sensor 1 minus the FoV
of sensor 2.
\item $A_{2}$ is the area that contains the FoV of sensor 2 minus the FoV
of sensor 1.
\item $A_{0}=\mathbb{R}^{n_{x}}\setminus\left(A_{1}\cup A_{2}\right)$ is
the union of the joint FoV and the area that is outside the FoV of
both sensors.
\end{itemize}
We then assign each Bernoulli component $f_{s}^{i}\left(\cdot\right)$
to one of these areas, and denote these as $f_{s}^{i,j}\left(\cdot\right)$
for $j\in\left\{ 0,1,2\right\} $ and $i=\left\{ 1,...,n_{s,j}\right\} $
where $n_{s,j}$ is the number of Bernoulli components in area $A_{j}$
for sensor $s$. How to perform this assignment will be explained
in Section \ref{subsec:Bernoulli-assignment}. 

Then, we write each PMB density as the union of three PMB densities
\begin{align}
f_{s}\left(X\right) & =\sum_{W^{0}\uplus W^{1}\uplus W^{2}=X}\prod_{j=0}^{2}f_{s}^{\mathrm{pmb},j}\left(W^{j}\right)\label{eq:Partitioned_PMB1}
\end{align}
where
\begin{align}
f_{s}^{\mathrm{pmb},j}\left(W^{i}\right) & =\sum_{X^{0}\uplus...\uplus X^{n_{s}}=X}f_{s}^{\mathrm{p},j}\left(X^{0}\right)\prod_{i=1}^{n_{s,j}}f_{s}^{i,j}\left(X^{i}\right)\label{eq:Partitioned_PMB2}
\end{align}
and the intensity of the PPP $f_{s}^{\mathrm{p},j}\left(\cdot\right)$
is $\lambda_{s}^{j}\left(x\right)=\lambda_{s}\left(x\right)\chi_{A_{j}}\left(x\right)$,
where $\chi_{A_{j}}\left(\cdot\right)$ is the indicator function
on $A_{j}$. That is, it is the intensity constrained to each of the
areas. It is important to note that writing the PMB (\ref{eq:PMB_density})
as (\ref{eq:Partitioned_PMB1})-(\ref{eq:Partitioned_PMB2}) is an
exact operation.

The next step is to use the GCI-PMB fusion rule to independently fuse
the two PMBs in each area with these settings:
\begin{itemize}
\item Fusion of $f_{1}^{\mathrm{pmb},0}\left(\cdot\right)$ with $f_{2}^{\mathrm{pmb},0}\left(\cdot\right)$
with $\omega\in\left(0,1\right)$, typically $\omega=1/2$, using
Theorem \ref{thm:Product-PMBs}, yielding a PMBM $q^{\mathrm{pmbm},0}\left(\cdot\right)$.
\item Fusion of $f_{1}^{\mathrm{pmb},1}\left(\cdot\right)$ with $f_{2}^{\mathrm{pmb},1}\left(\cdot\right)$
with $\omega=1$, resulting in $f_{1}^{\mathrm{pmb},1}\left(\cdot\right)$.
\item Fusion of $f_{1}^{\mathrm{pmb},2}\left(\cdot\right)$ with $f_{2}^{\mathrm{pmb},2}\left(\cdot\right)$
with $\omega=0$, resulting in $f_{2}^{\mathrm{pmb},2}\left(\cdot\right)$.
\end{itemize}
That is, we set $\omega$ to either 0 or 1 if only one of the sensors
clearly has more accurate information than the other in the associated
area. The resulting fused density is
\begin{align}
q\left(X\right) & =\sum_{W^{0}\uplus W^{1}\uplus W^{2}=X}q^{\mathrm{pmbm},0}\left(W^{0}\right)\nonumber \\
 & \quad\times f_{1}^{\mathrm{pmb},1}\left(W^{1}\right)f_{2}^{\mathrm{pmb},2}\left(W^{2}\right)\label{eq:fused_partitioned_PMBM}
\end{align}
It should be noted that this density is a PMBM. It can be written
in the PMBM standard form \cite{Angel18_b} by first noticing that
its PPP intensity corresponds to the sum of the intensities of the
PPPs of $q^{\mathrm{pmbm},0}\left(\cdot\right)$, $f_{1}^{\mathrm{pmb},1}\left(\cdot\right)$
and $f_{2}^{\mathrm{pmb},2}\left(\cdot\right)$, and then appending
the Bernoulli components in $f_{1}^{\mathrm{pmb},1}\left(\cdot\right)$
and $f_{2}^{\mathrm{pmb},2}\left(\cdot\right)$ to all the multi-Bernoulli
components in the mixture.

\subsection{Bernoulli assignment\label{subsec:Bernoulli-assignment}}

In this section, we explain how to assign the Bernoulli components
of sensors 1 and 2 to areas $A_{0}$, $A_{1}$ and $A_{2}$. A simple
option would be to assign each Bernoulli to the area that contains
most of its probability. However, this can lead to missed object problems
at the border of the areas if two Bernoulli components that represent
the same object are associated to different areas. 

A better alternative is cluster Bernoulli components of both sensors
by similarity and assign all the Bernoulli components in each cluster
to the same area. To do this, we first compute a distance between
the Bernoulli components in one sensor and the other sensor, e.g.,
using the square Mahalanobis distance in (\ref{eq:Mahalanobis}).
If two Bernoulli components are closer than a threshold, we consider
they can be assigned. Then, the clusters are given by pairs of subsets
of the Bernoulli components in sensor 1 and in sensor 2 in which there
can be assignments. These can be computed by first obtaining the adjacency
matrix of the (unweighted) bipartite graph formed by the Bernoulli
components, where there is an edge if the Bernoulli components can
be assigned. Then, the clusters are the disjoint components in this
graph, which can be computed via the reverse Cuthill-McKee algorithm
\cite{George_book81}.

\section{Simulations\label{sec:Simulations}}

In this section, we evaluate the performance of the GCI PMB fusion
rule for distributed MOF via numerical simulations. We first consider
a scenario with two agents. Then we consider a scenario with four
agents with limited FoVs. We compare the following distributed PMBM-PMB\footnote{Matlab code of the proposed algorithms is available at https://github.com/Agarciafernandez/MTT.}
filter variants, implemented with both GCI and AA fusion.
\begin{itemize}
\item Distributed PMBM (DPMBM) filter: each agent runs a PMBM filter with
a variational PMB projection with the most likely assignment \cite[Sec. V.B]{Xia22}
before applying the fusion step. 
\item DPMB-TO (distributed PMB track oriented) filter: each agent runs a
PMB filter performing a track-oriented PMB projection \cite{Williams15b,Angel20_e}
after each update and fusion step.
\item DPMB-V: each agent runs a PMB filter performing a variational PMB
projection with the most likely assignment \cite[Sec. V.B]{Xia22}
after each update and after the fusion rule.
\item DPMB-GNN: each agent runs a PMB filter with the best global nearest
neighbour (GNN) hypothesis for each update and fusion.
\item DPMB-KUB: each agent runs a track-oriented PMB filter, and uses the
KLD upper bound (KUB) fusion rule in \cite[Sec. V.B]{Frohle20}. This
method is only valid for GCI fusion.
\item Centralised PMBM: sequential update with the measurements of each
agent.
\end{itemize}

\subsection{Scenario 1}

The scenario has two agents tracking point objects, and the fusion
rule is performed every $N_{f}$ time steps. The parameters of the
algorithms are the following. The PMBM/PMB implementation follows
the one in \cite{Angel18_b}. The filters propagate a PPP with maximum
30 components, pruning threshold $10^{-5}$ and merging threshold
0.1. The threshold for pruning multi-Bernoulli densities is $10^{-4}$.
For data association, we use gating with threshold 20 and Murty's
algorithm with 200 hypotheses. Bernoulli components with a probability
of existence lower than $10^{-5}$ are discarded. PMBM pruning and
merging are performed before and after each VPMB projection to lower
computational cost. The VPMB projection is done with a maximum of
10 iterations, and a convergence threshold 0.1.

The GCI fusion rule has parameter $\omega=1/2$, gating with threshold
$\Gamma_{f}=20$ and Murty's algorithm with 200 hypotheses. For comparison
purposes, we consider the AA fusion rule for PMB/PMBM filtering \cite{Li23b}.
The AA fusion rule has parameter $\omega=1/2$ and we have implemented
a Bernoulli-to-Bernoulli association method based on the Mahalanobis
distance (other options are possible \cite{Li20}). That is, we start
with all Bernoulli components being unassigned. Then, we go through
all Bernoulli components of agent 1 and compute the Mahalanobis distance
with all means of the unassigned Bernoulli components in agent 2.
If the minimum square distance is lower than a threshold $\Gamma_{aa}=5$,
these Bernoulli components are assigned, and belong to the same track
in a PMBM density structure \cite{Williams15b}.

\begin{figure}
\begin{centering}
\includegraphics[scale=0.6]{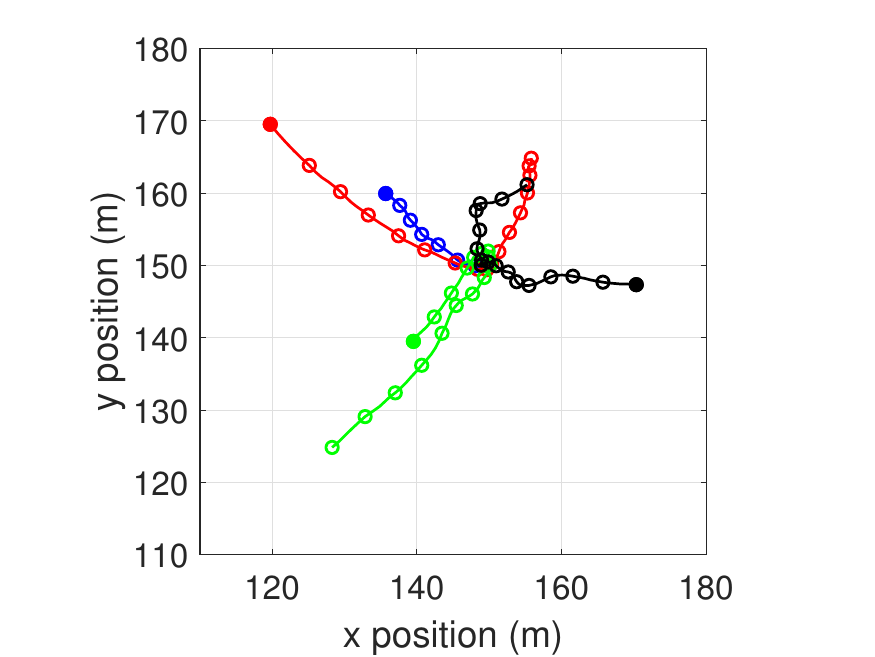}
\par\end{centering}
\caption{\label{fig:Scenario1}Scenario 1, similar to \cite{Angel18_b}. There
are four objects that get in close proximity in the middle of the
simulation. All objects are born at the beginning of the simulation
and the blue one dies at time step 40. The object positions every
10 time steps are marked with a circle, and the initial positions
with filled circles.}

\end{figure}

The single-object state is $x=\left[p_{x},\dot{p}_{x},p_{y},\dot{p}_{y}\right]^{T}$
containing its position and velocity. Each object moves with a nearly-constant
velocity model with sampling interval $\tau=1$ and noise covariance
parameter 0.01 \cite{Angel18_b}. The probability of survival is 0.99.
The birth model is a PPP with Gaussian intensity with mean $\left[100,0,100,0\right]^{T}$
and covariance matrix $\mathrm{diag}\left(\left[150^{2},1,150^{2},1\right]\right)$.
The expected number of new born targets is 3 at time step 1 and 0.005
in the rest of the time steps. There are 81 time steps in the simulations
and the ground truth object trajectories are shown in Figure \ref{fig:Scenario1}.

The agents measure the object positions with additive zero-mean Gaussian
noise with covariance matrix $R=4I$, and a probability of detection
$p^{D}=0.9$. Clutter is PPP distributed with clutter intensity is
$\lambda^{C}\left(z\right)=\overline{\lambda}^{C}u_{A}\left(z\right)$
where $u_{A}\left(z\right)$ is a uniform density in $A=\left[0,300\right]\times\left[0,300\right]$
and $\overline{\lambda}^{C}=10$.

\begin{figure}
\begin{centering}
\includegraphics[scale=0.3]{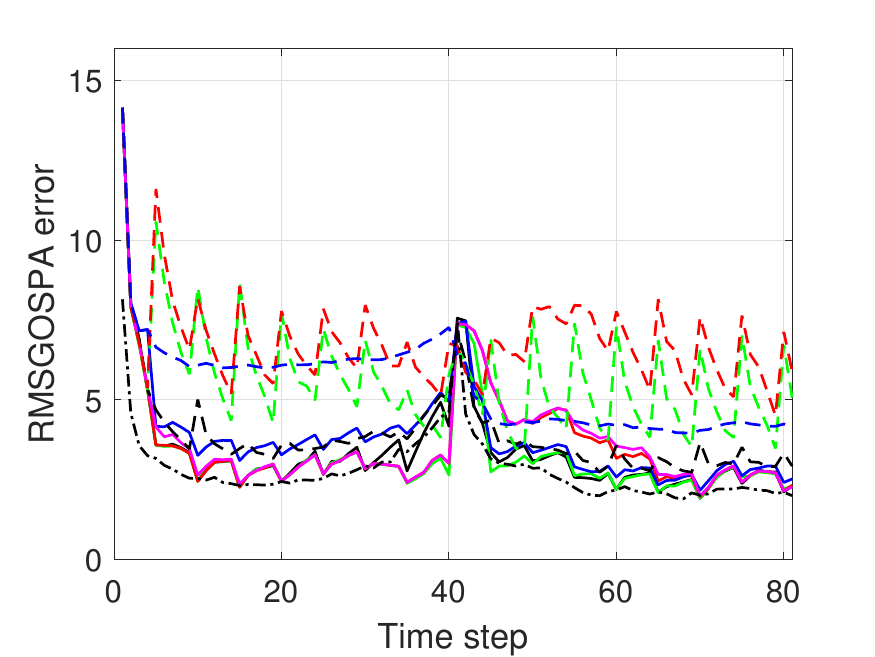}\includegraphics[scale=0.3]{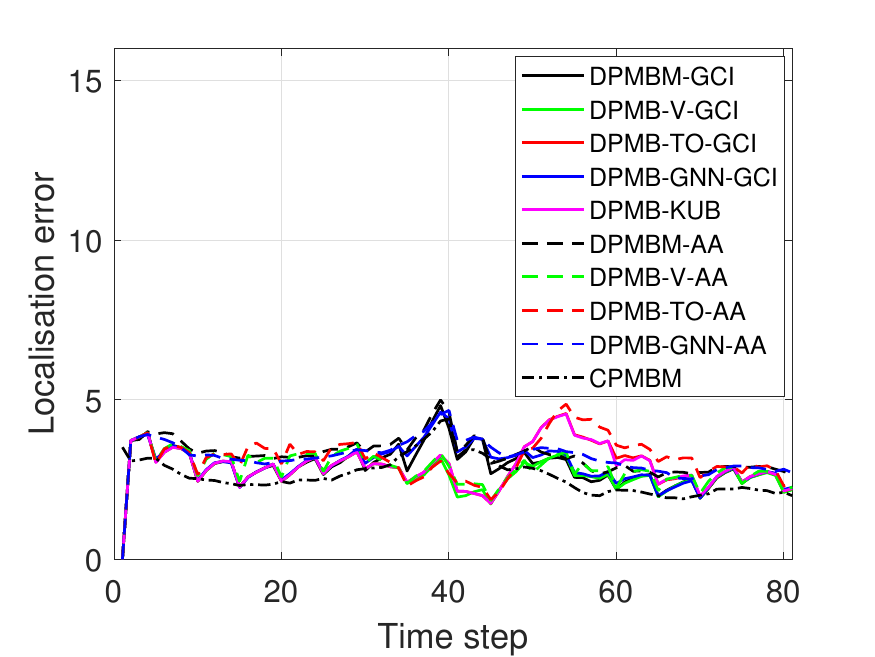}
\par\end{centering}
\begin{centering}
\includegraphics[scale=0.3]{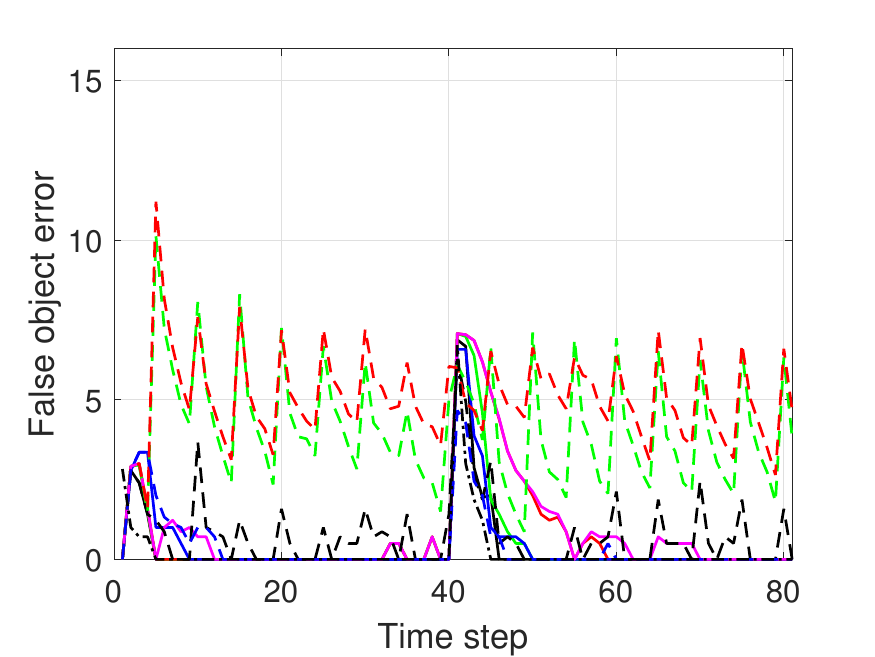}\includegraphics[scale=0.3]{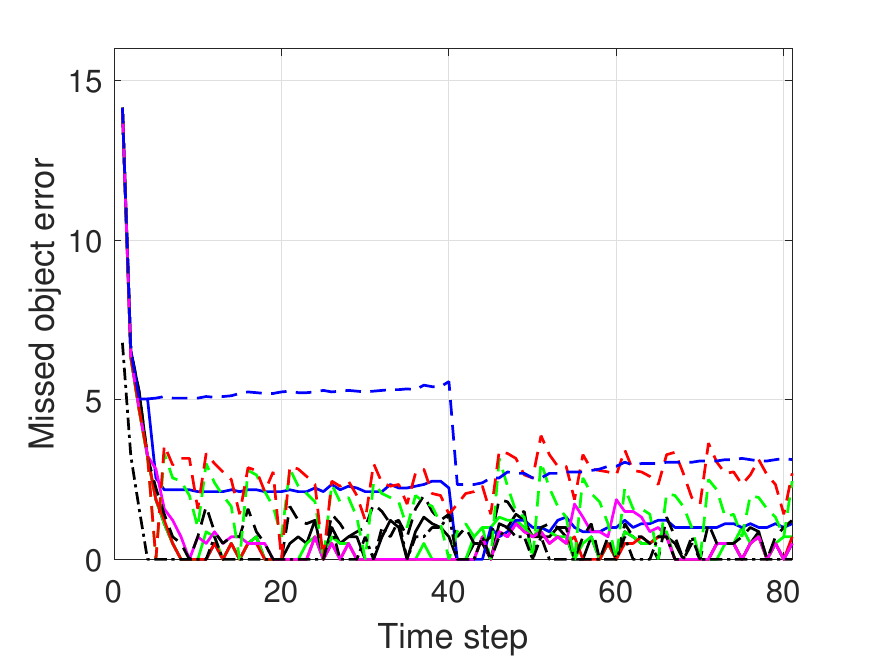}
\par\end{centering}
\caption{\label{fig:RMS-GOSPA-errors_Sce1}RMS-GOSPA errors at each time step
for $N_{f}=5$ (Scenario 1). When the agents perform information fusion
the error decreases. Among the distributed filters, DPMB-V-GCI and
DPMBM-GCI have the best performance.}
\end{figure}

MOF performance is evaluated via Monte Carlo simulation with 100 runs.
For each agent, we calculate the error in the estimated set of object
positions using generalised optimal subpattern assignment (GOSPA)
metric with parameters $\alpha=2$, $p=2$ and $c=10$ \cite{Rahmathullah17}.
We first analyse the case $N_{f}=5$, which means the agents fuse
their information every 5 time steps. The root mean square GOSPA (RMS-GOSPA)
errors at each time step, considering both agents, are shown in Figure
\ref{fig:RMS-GOSPA-errors_Sce1}. This figure also shows the decomposition
of the GOSPA metric into localisation cost, missed object cost and
false object cost. As expected, the best performance is obtained by
the centralised filter, the CPMBM. Among the distributed filters,
the DPMB-V-GCI and DPMBM-GCI are the best performing filters. DPMB-TO-GCI
performs well before the objects get in close proximity but then the
error is higher than the best performing filters due to an increase
in false objects. AA filter variants also perform well but they show
a higher number of false objects, which increases their GOSPA errors.
The GNN versions of the filters perform worse due to a higher number
of missed objects. This is expected since the GNN filters only take
into account the best global hypothesis, both in the update and in
the information fusion step. We can see that every 5 time steps there
is a reduction in the error in the distributed filters. This corresponds
to the time steps when there is a data fusion step. 

We also show the RMS-GOSPA errors across all time steps for $N_{f}\in\{1,5,10\}$
in Table \ref{tab:RMS-GOSPA-errors-all-times}. As expected, the best
performing filter is the CPMBM filter. The best performing distributed
filters are DPMBM-GCI and DPMB-V. For these filters (and also DPMB-TO-GCI)
lowering $N_{f}$ increases performance, which is reasonable since
there are more fusion steps. DPMB-TO-GCI exhibits higher error due
to some track coalescence when targets gets in close proximity and
then separate. DPMB-GNN-GCI has a higher error due to missed targets.
Generally, AA performs worse than GCI in this scenario. We should
recall that Bernoulli-to-Bernoulli association (with the possibility
of leaving Bernoulli components unassigned) is a natural part of the
GCI fusion rule, as shown in Theorem \ref{thm:Product-PMBs}. However,
AA does not have a Bernoulli-to-Bernoulli association step as a fundamental
step of the algorithm, but it must be performed via an external procedure
to achieve good performance. It is likely that this difference contributes
to the better performance of GCI in these simulations. DPMB-KUB works
worse than DPMBM, DPMB-V and DPMB-TO with the GCI rule, but it works
better than DPMB-GNN and the AA variants (except DPMBM-AA).

\begin{table}
\caption{\label{tab:RMS-GOSPA-errors-all-times}RMS-GOSPA errors across all
time steps}

\centering{}%
\begin{tabular}{c|ccc|ccc}
\hline 
Fusion rule &
\multicolumn{3}{c|}{GCI} &
\multicolumn{3}{c}{AA}\tabularnewline
\hline 
$N_{f}$ &
1 &
5 &
10 &
1 &
5 &
10\tabularnewline
\hline 
DPMBM &
\uline{3.34} &
3.77 &
3.96 &
4.15 &
4.16 &
4.09\tabularnewline
DPMB-V &
3.45 &
\uline{3.69} &
\uline{3.87} &
5.40 &
5.90 &
5.22\tabularnewline
DPMB-TO &
3.77 &
4.02 &
4.16 &
8.95 &
6.93 &
6.07\tabularnewline
DPMB-GNN &
5.20 &
4.17 &
4.49 &
5.63 &
5.64 &
5.64\tabularnewline
DPMB-KUB &
4.88 &
4.07 &
4.20 &
- &
- &
-\tabularnewline
\hline 
CPMBM &
\multicolumn{6}{c}{2.99}\tabularnewline
\hline 
\end{tabular}
\end{table}

\subsection{Scenario 2 }

\begin{figure}
\begin{centering}
\includegraphics[scale=0.6]{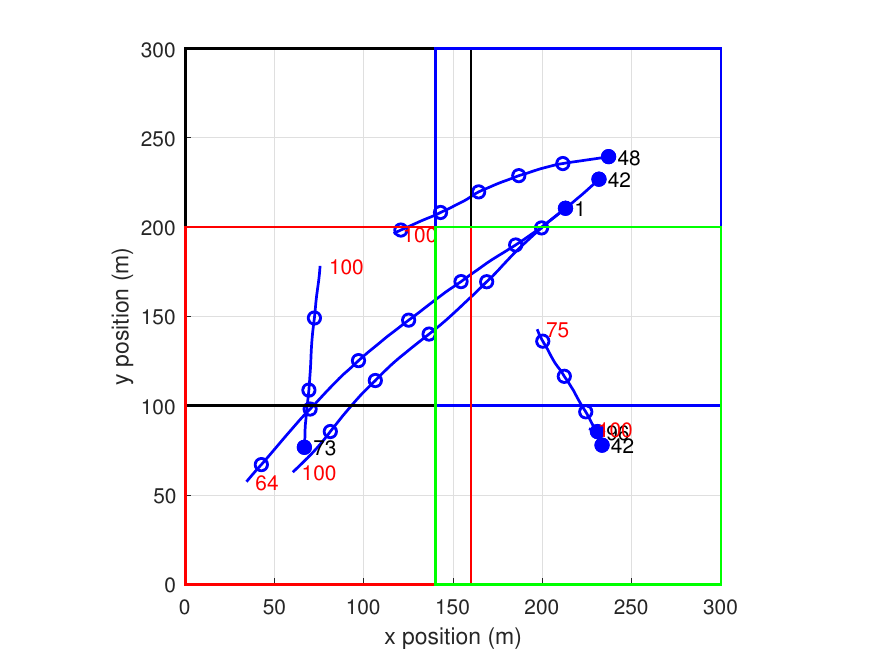}
\par\end{centering}
\caption{\label{fig:Scenario2}Scenario 2 considering 4 agents with limited
FoVs and 100 time steps. The FoV of each agent is a rectangle shown
with a different colour. There are six objects, whose positions are
marked every 10 time steps with a circle, and the initial positions
with filled circles. The initial time step of each object is written
in black next to its initial position. The final time step is written
in red next to its final position.}
\end{figure}

This experiment is based on the scenario shown in Figure \ref{fig:Scenario2},
which contains 4 agents with limited FoVs. The GCI coefficients to
perform the fusion rule sequentially are those provided in \cite[Table I]{Gunay16}.
All the algorithms have been implemented with the partitioning approach
described in Section \ref{sec:Application-limited_Fov}, considering
the union of the FoVs of the agents for which we have already performed
the fusion, as was done in \cite[Alg. 1]{Wang22b}. All agents fuse
their information every $N_{f}=5$ time steps. We have not managed
to get DPMB-UBK working properly in this scenario so its results are
omitted. 

In this scenario, the filters consider a maximum of 100 global hypotheses,
and a maximum of 10 PPP components. The GCI-fusion rule also uses
100 global hypotheses with gating threshold 20. Merging of PPP components
is performed in their square Mahalanobis distance is smaller than
4. After each PMB projection, we perform recycling with threshold
0.01. We also perform merging of two Bernoulli components if their
square Mahalanobis distance is smaller than 4, and the probability
that there are two objects after the merging is smaller than 0.01.
The probability of existence of the merged Bernoulli components is
their sum, which matches the PHD, capped to one. 

The single-object state, dynamic model and probability of survival
are the same as in the previous section. The birth model is a PPP
with intensity
\begin{align}
\lambda^{B}\left(x\right) & =\sum_{q=1}^{4}w_{k}^{b,q}\mathcal{N}\left(x;\overline{x}^{b,q},P^{b,q}\right)
\end{align}
where $w_{0}^{b,q}=0.75$, $w_{k}^{b,q}=0.01$ for $k>0$, $P^{b,q}=\mathrm{diag}\left(\left[10^{2},1,10^{2},1\right]\right)$,
$\overline{x}^{b,1}=\left[75,2,225,-2\right]^{T}$, $\overline{x}^{b,2}=\left[225,-2,225,-2\right]^{T}$,
$\overline{x}^{b,3}=\left[75,1,75,2\right]^{T}$, and $\overline{x}^{b,4}=\left[225,-1,75,2\right]^{T}$.
There is a Gaussian birth component in the FoV of each sensors. 

The agents measure the object positions with zero-mean Gaussian noise
with covariance $R=4I$. The probability of detection of sensor $s$,
with field of view $\mathrm{FoV}_{s}\subset\mathbb{R}^{n_{x}}$, is
\begin{align}
p_{s}^{D}\left(x\right) & =\begin{cases}
p^{D} & x\in\mathrm{FoV}_{s}\\
0 & \mathrm{otherwise}
\end{cases}
\end{align}
with $p^{D}=0.9$. Clutter for each sensor is a PPP with uniform intensity
in the FoV of each sensor (in the single-measurement space $\mathbb{R}^{n_{z}}$)
with $\overline{\lambda}^{C}=10$. For each single-object Gaussian
update, all filters approximate the state-dependent $p_{s}^{D}\left(\cdot\right)$
as a constant given by its expected value. This is approximately computed
by Monte Carlo integration, sampling from the prior, with 100 samples. 

\begin{figure}
\begin{centering}
\includegraphics[scale=0.3]{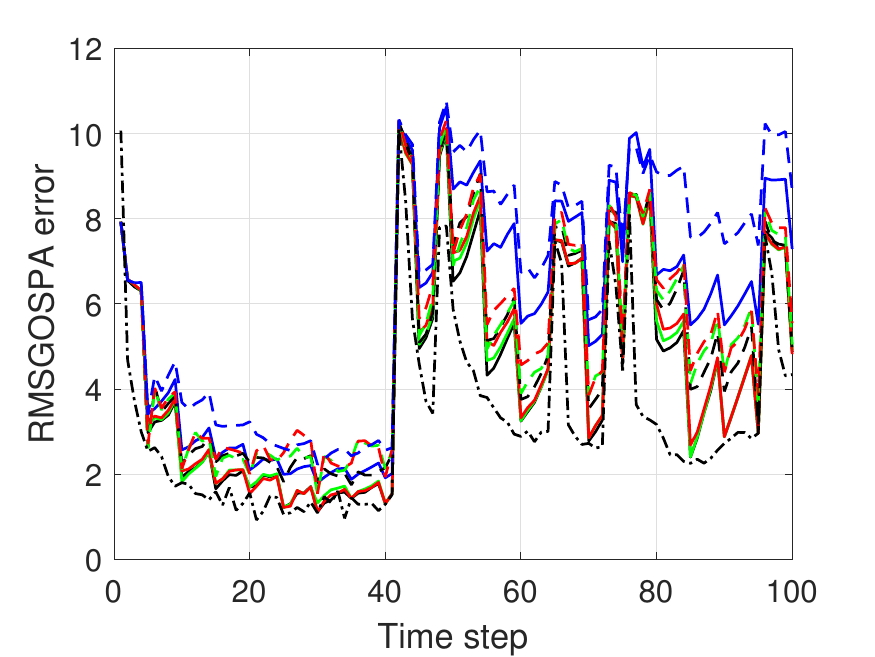}\includegraphics[scale=0.3]{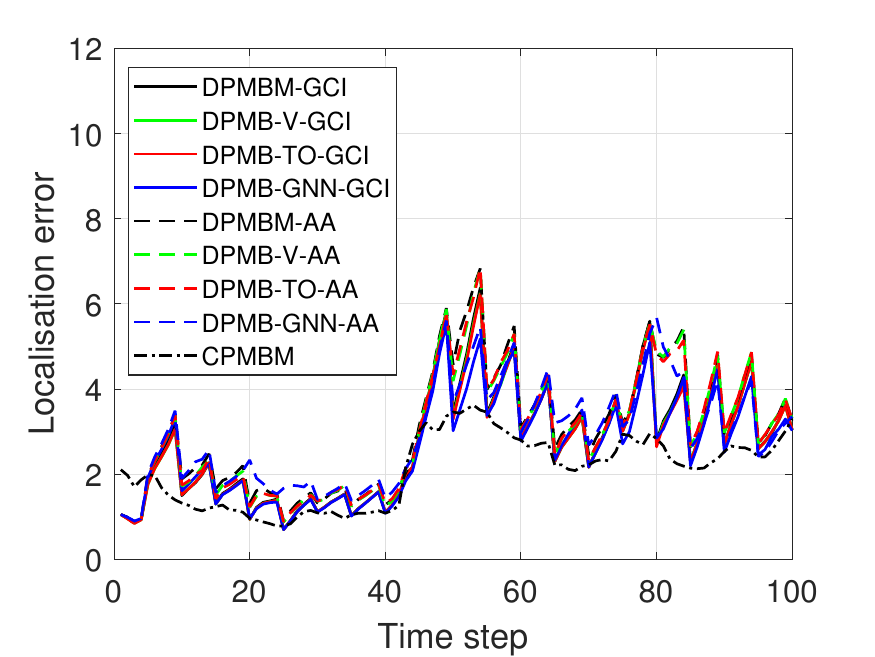}
\par\end{centering}
\begin{centering}
\includegraphics[scale=0.3]{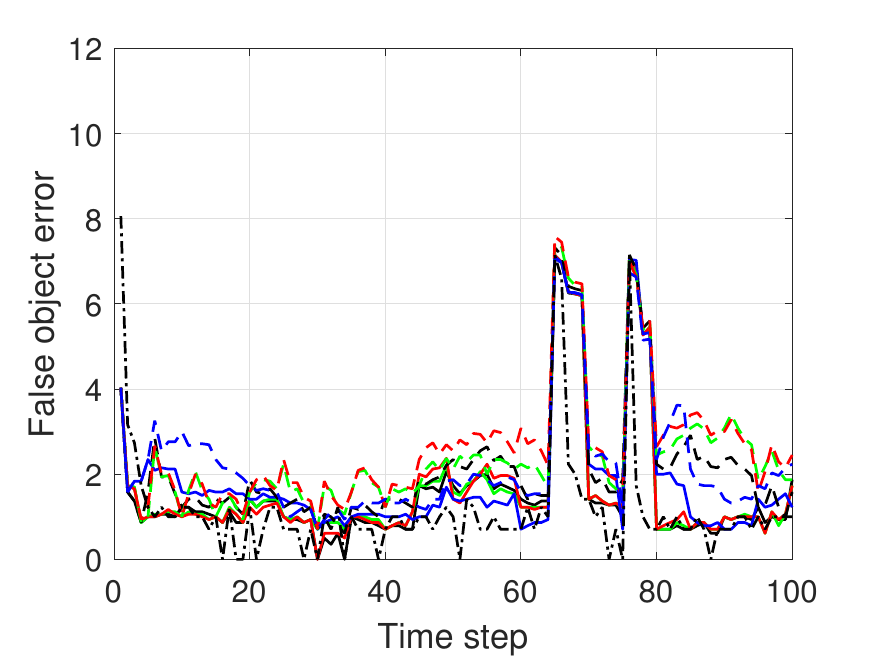}\includegraphics[scale=0.3]{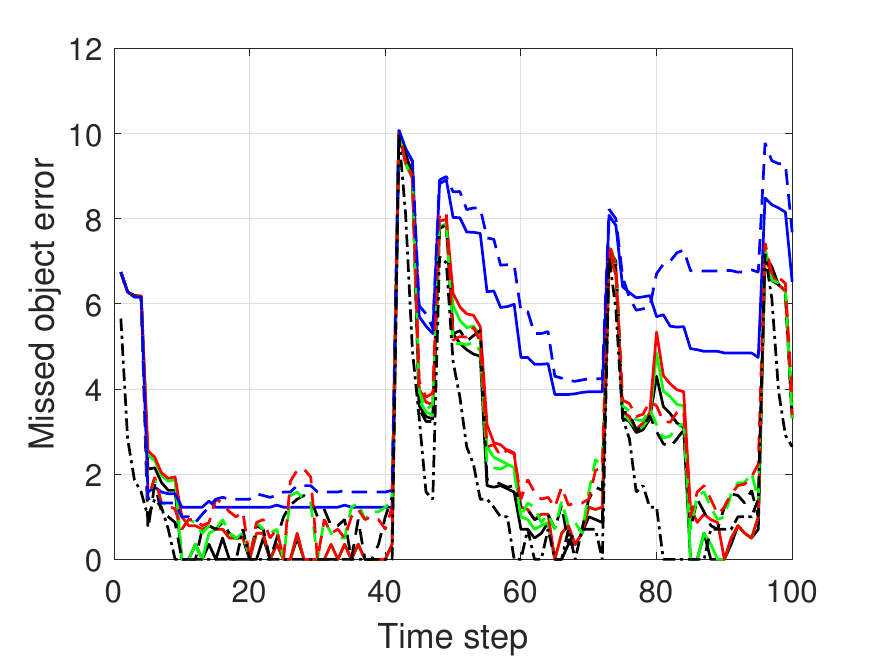}
\par\end{centering}
\caption{\label{fig:RMS-GOSPA-errors_Sce2}RMS-GOSPA errors at each time step
for $N_{f}=5$ (Scenario 2). Among the distributed filters, DPMB-V-GCI
and DPMBM-GCI have the best performance. }

\end{figure}

At each time step, all agents estimate the set of objects. The root
mean square GOSPA (RMS-GOSPA) errors and their decompositions at each
time step, considering all agents, using Monte Carlo simulation with
100 runs is shown in Figure \ref{fig:RMS-GOSPA-errors_Sce2}. The
centralised PMBM filter provides the lowest errors. After 5 time steps,
the distributed filters improve performance, as the exchange of information
happens every 5 time steps. DPMBM-GCI and DPMB-V-GCI have the best
performance among the distributed filters. We can see that a main
difference between the centralised solution and these distributed
filters is in the localisation error, which is higher in the distributed
filters. Nevertheless, when agents exchange their information, the
localisation error is close to the centralised implementation. False
and missed objects tend to linger more in the distributed filters,
also due to the delay in the exchange of information. The filter with
worse performance is DPMB-GNN-AA, especially due to an increase in
the missed object error. Overall, the filters with the GCI rule perform
better than those with AA.

\section{Conclusions\label{sec:Conclusions}}

This paper has derived an approximate GCI fusion rule for PMB densities.
The only approximation involved is to approximate the power of a PMB
by an unnormalised PMB, which is an upper bound. The rest of the GCI
fusion rule is in closed form and yields a PMBM that accounts for
all possible assignments of Bernoulli components among the two PMB
densities. The proposed fusion rule is applicable to any PMB filter,
for instance, (track-oriented) PMB filter \cite{Williams15b}, (measurement-oriented)
PMB filter \cite{Williams15b}, variational PMB filter \cite{Williams15},
PMB filters for extended object tracking \cite{Xia22} or PMB filters
for co-existing point and extended object tracking \cite{Angel21}.
The fusion rule can also be applied to any PMBM filter by performing
a PMB projection beforehand. The fusion rule has also been extended
to deal with sensors with limited FoVs by applying different fusion
coefficients in different areas.

Future work includes the application of this fusion rule to large
networks via consensus \cite{Battistelli13} and the extension of
this fusion rule to PMBMs. Another line of future work is the distributed
recovery of trajectory information by each agent, for instance, using
the sequence of filtered PMB densities and backward simulation \cite{Xia22b}.

\bibliographystyle{IEEEtran}
\bibliography{13C__Trabajo_laptop_Referencias_Referencias}

\cleardoublepage{}

{\LARGE Supplemental material: ``Distributed Poisson multi-Bernoulli
filtering via generalised covariance intersection''}{\LARGE\par}

\appendices{}

\section{\label{sec:Proof_power_PMB}}

In this appendix, we prove Lemma \ref{lem:Omega_power_PMB}. From
Approximation (\ref{eq:approximation_power_PMB}), the $\omega$-power
of the PPP density is
\begin{align}
\left(f_{1}^{\mathrm{p}}\left(X^{0}\right)\right)^{\omega} & =e^{-\omega\int\lambda_{1}\left(x\right)dx}\left[\lambda_{1}^{\omega}\right]^{X}\nonumber \\
 & =\frac{e^{-\omega\int\lambda_{1}\left(x\right)dx}}{e^{-\int\lambda_{1}^{\omega}\left(x\right)dx}}e^{-\int\lambda_{1}^{\omega}\left(x\right)dx}\left[\lambda_{1}^{\omega}\right]^{X}\nonumber \\
 & =\frac{e^{-\omega\int\lambda_{1}\left(x\right)dx}}{e^{-\int\lambda_{1}^{\omega}\left(x\right)dx}}q_{1}^{\mathrm{p}}\left(X\right).\label{eq:f_Poisson_power}
\end{align}

On the other hand, the $\omega$-power of the $i$-th Bernoulli density
is

\begin{align}
\left(f_{1}^{i}\left(X^{i}\right)\right)^{\omega} & =\begin{cases}
\left(1-r_{s}^{i}\right)^{\omega} & X=\emptyset\\
\left(r_{s}^{i}\right)^{\omega}\left(p_{s}^{i}\left(x\right)\right)^{\omega} & X=\left\{ x\right\} \\
0 & \mathrm{otherwise}.
\end{cases}\nonumber \\
 & =\begin{cases}
\left(1-r_{s}^{i}\right)^{\omega} & X=\emptyset\\
\left(r_{s}^{i}\right)^{\omega}p_{q,1}^{i}\left(x\right)\int\left(p_{1}^{i}\left(x\right)\right)^{\omega}dx & X=\left\{ x\right\} \\
0 & \mathrm{otherwise}.
\end{cases}
\end{align}
The above equation can be written as
\begin{align}
\left(f_{1}^{i}\left(X^{i}\right)\right)^{\omega} & =\frac{\left(1-r_{s}^{i}\right)^{\omega}+\left(r_{s}^{i}\right)^{\omega}\int\left(p_{1}^{i}\left(x\right)\right)^{\omega}dx}{\left(1-r_{s}^{i}\right)^{\omega}+\left(r_{s}^{i}\right)^{\omega}\int\left(p_{1}^{i}\left(x\right)\right)^{\omega}dx}\nonumber \\
 & \times\begin{cases}
\left(1-r_{s}^{i}\right)^{\omega} & X=\emptyset\\
\left(r_{s}^{i}\right)^{\omega}p_{q,1}^{i}\left(x\right)\int\left(p_{1}^{i}\left(x\right)\right)^{\omega}dx & X=\left\{ x\right\} \\
0 & \mathrm{otherwise}.
\end{cases}\nonumber \\
 & =\left[\left(1-r_{s}^{i}\right)^{\omega}+\left(r_{s}^{i}\right)^{\omega}\int\left(p_{1}^{i}\left(x\right)\right)^{\omega}dx\right]q_{1}^{i}\left(X\right).\label{eq:f_Bernoulli_power}
\end{align}

Substituting (\ref{eq:f_Poisson_power}) and (\ref{eq:f_Bernoulli_power})
into (\ref{eq:approximation_power_PMB}) results in (\ref{eq:approximation_power_PMB2})
proving Lemma \ref{lem:Omega_power_PMB}.

\section{\label{sec:Proof_Theorem1}}

In this appendix we prove Theorem \ref{thm:Product-PMBs} on the product
of PMBs. We can write (\ref{eq:normalised_product_PMBs}) as
\begin{align}
q\left(X\right) & \propto\sum_{X^{0}\uplus...\uplus X^{n_{1}}=X}\left[q_{1}^{\mathrm{p}}\left(X^{0}\right)\prod_{i=1}^{n_{1}}q_{1}^{i}\left(X^{i}\right)\right.\nonumber \\
 & \,\times\left.\sum_{Y^{0}\uplus...\uplus Y^{n_{2}}=X^{0}\uplus...\uplus X^{n_{1}}}q_{2}^{\mathrm{p}}\left(Y^{0}\right)\prod_{j=1}^{n_{2}}q_{2}^{j}\left(Y^{j}\right)\right].\label{eq:product_PMBs_append}
\end{align}
Let us focus on the inner sum. As $Y^{1},...,Y^{n_{2}}$ and $X^{1}...X^{n_{1}}$
can have at maximum one element (otherwise the corresponding Bernoulli
density is zero), we can assign the sets in $Y^{1},...,Y^{n_{2}}$
to the sets in $X^{1}...X^{n_{1}}$. The sets in $Y^{1},...,Y^{n_{2}}$
that remain unassigned will be associated with $X^{0}$. The sets
in $X^{1}...X^{n_{1}}$ that remain unassigned will be associated
to $Y^{0}$. The sum can then be written as a sum over all possible
assignments between $Y^{1},...,Y^{n_{2}}$ and $X^{1}...X^{n_{1}}$,
which we denote by the assignment set $\gamma\in\Gamma$, see the
paragraph before Theorem \ref{thm:Product-PMBs} for its definition.

\textcolor{red}{}

For the assignment $\gamma=\left\{ \gamma_{1},...,\gamma_{n}\right\} $,
where $n\in\left\{ 1,...,\min(n_{1},n_{2})\right\} $ being $\gamma_{i}=\left(\gamma_{i}(1),\gamma_{i}(2)\right)$,
we have that
\begin{align}
 & Y^{\gamma_{1}(2)}=X^{\gamma_{1}(1)},...,Y^{\gamma_{n}(2)}=X^{\gamma_{n}(1)}.\label{eq:assignment_Y_X}
\end{align}

In addition, for this $\gamma$, we denote the indices of the unassigned
sets in $X^{1}$,...,$X^{n_{1}}$ as $\left\{ \gamma_{1}^{u,1},...,\gamma_{n_{1}-n}^{u,1}\right\} $.
Similarly, the unassigned sets in $Y^{1},...,Y^{n_{2}}$ have indices
$\left\{ \gamma_{1}^{u,2},...,\gamma_{n_{2}-n}^{u,2}\right\} $. In
the inner convolution sum in (\ref{eq:product_PMBs_append}), there
is the following equality
\begin{align}
Y^{0}\uplus Y^{1}\uplus...\uplus Y^{n_{2}} & =X^{0}\uplus X^{1}\uplus...\uplus X^{n_{1}}.
\end{align}
For the assignment $\gamma$ using (\ref{eq:assignment_Y_X}), this
equality becomes
\begin{align}
 & Y^{0}\uplus X^{\gamma_{1}(1)}\uplus...\uplus X^{\gamma_{n}(1)}\uplus Y^{\gamma_{1}^{u,2}}\uplus...\uplus Y^{\gamma_{n_{2}-n}^{u,2}}\nonumber \\
 & =X^{0}\uplus X^{\gamma_{1}(1)}\uplus...\uplus X^{\gamma_{n}(1)}\uplus X^{\gamma_{1}^{u,1}}\uplus...\uplus X^{\gamma_{n_{1}-n}^{u,1}}.
\end{align}
This implies that
\begin{align*}
Y^{0}\uplus Y^{\gamma_{1}^{u,2}}\uplus...\uplus Y^{\gamma_{n_{2}-n}^{u,2}} & =X^{0}\uplus X^{\gamma_{1}^{u,1}}\uplus...\uplus X^{\gamma_{n_{1}-n}^{u,1}}
\end{align*}
and
\begin{align}
Y^{0} & =X^{0}\uplus X^{\gamma_{1}^{u,1}}\uplus...\uplus X^{\gamma_{n_{1}-n}^{u,1}}\setminus\left(Y^{\gamma_{1}^{u,2}}\uplus...\uplus Y^{\gamma_{n_{2}-n}^{u,2}}\right).\label{eq:Y0_append1}
\end{align}

We introduce an additional set $U$, representing objects that are
unassigned in both densities, such that
\begin{align}
X^{0} & =U\uplus Y^{\gamma_{1}^{u,2}}\uplus...\uplus Y^{\gamma_{n_{2}-n}^{u,2}}\label{eq:X_0}
\end{align}
and therefore (\ref{eq:Y0_append1}) becomes
\begin{align}
Y^{0} & =U\uplus X^{\gamma_{1}^{u,1}}\uplus...\uplus X^{\gamma_{n_{1}-n}^{u,1}}.\label{eq:Y0}
\end{align}
Summing over all assignment sets and substituting these results into
(\ref{eq:product_PMBs_append}) yields

\begin{align}
 & q\left(X\right)\nonumber \\
 & \propto\sum_{n=0}^{\min(n_{1},n_{2})}\sum_{\left\{ \gamma_{1},...,\gamma_{n}\right\} \in\Gamma}\sum_{Y^{\gamma_{1}^{u,2}}\uplus...\uplus Y^{\gamma_{n_{2}-n}^{u,2}}\uplus U\uplus X^{1}...\uplus X^{n_{1}}=X}\nonumber \\
 & \times\prod_{i=1}^{n}q_{1}^{\gamma_{i}(1)}\left(X^{\gamma_{1}(1)}\right)q_{2}^{\gamma_{i}(2)}\left(X^{\gamma_{1}(1)}\right)\nonumber \\
 & \times q_{2}^{\mathrm{p}}\left(U\uplus X^{\gamma_{1}^{u,1}}\uplus...\uplus X^{\gamma_{n_{1}-n}^{u,1}}\right)\prod_{i=1}^{n_{1}-n}q_{1}^{\gamma_{i}^{u,1}}\left(X^{\gamma_{i}^{u,1}}\right)\nonumber \\
 & \times q_{1}^{\mathrm{p}}\left(U\uplus Y^{\gamma_{1}^{u,2}}\uplus...\uplus Y^{\gamma_{n_{2}-n}^{u,2}}\right)\prod_{i=1}^{n_{2}-n}q_{2}^{\gamma_{i}^{u,2}}\left(Y^{\gamma_{i}^{u,2}}\right)
\end{align}
where the second line is the consequence of applying (\ref{eq:assignment_Y_X}),
the third line is the consequence of applying (\ref{eq:Y0}) and keeping
the unassigned Bernoulli components in density 1, the fourth line
results from (\ref{eq:X_0}) and keeping the unassigned Bernoulli
components in density 2.

Making use of the fact that
\begin{align*}
q_{2}^{\mathrm{p}}\left(U\uplus X^{\gamma_{1}^{u,1}}\uplus...\uplus X^{\gamma_{n_{1}-n}^{u,1}}\right) & \propto q_{2}^{\mathrm{p}}\left(U\right)\prod_{i=1}^{n_{1}-n}q_{2}^{\mathrm{p}}\left(X^{\gamma_{i}^{u,1}}\right)\\
q_{1}^{\mathrm{p}}\left(U\uplus Y^{\gamma_{1}^{u,2}}\uplus...\uplus Y^{\gamma_{n_{2}-n}^{u,2}}\right) & \propto q_{1}^{\mathrm{p}}\left(U\right)\prod_{i=1}^{n_{2}-n}q_{1}^{\mathrm{p}}\left(Y^{\gamma_{i}^{u,2}}\right),
\end{align*}
we obtain
\begin{align}
 & q\left(X\right)\nonumber \\
 & \text{\ensuremath{\propto}}\sum_{n=0}^{\min(n_{1},n_{2})}\sum_{\left\{ \gamma_{1},...,\gamma_{n}\right\} \in\Gamma}\nonumber \\
 & \sum_{U\uplus X^{1}\uplus...\uplus X^{n_{1}}\uplus Y^{\gamma_{1}^{u,2}}\uplus...\uplus Y^{\gamma_{n_{2}-n}^{u,2}}=X}\nonumber \\
 & \times\prod_{i=1}^{n}q_{1}^{\gamma_{i}(1)}\left(X^{\gamma_{1}(1)}\right)q_{2}^{\gamma_{i}(2)}\left(X^{\gamma_{1}(1)}\right)\nonumber \\
 & \times q_{2}^{\mathrm{p}}\left(U\right)q_{1}^{\mathrm{p}}\left(U\right)\nonumber \\
 & \times\prod_{i=1}^{n_{1}-n}\left[f_{1}^{\gamma_{i}^{u,1}}\left(X^{\gamma_{i}^{u,1}}\right)q_{2}^{\mathrm{p}}\left(X^{\gamma_{i}^{u,1}}\right)\right]\nonumber \\
 & \times\prod_{i=1}^{n_{2}-n}\left[q_{1}^{\mathrm{p}}\left(Y^{\gamma_{i}^{u,2}}\right)q_{2}^{\gamma_{i}^{u,2}}\left(Y^{\gamma_{i}^{u,2}}\right)\right].
\end{align}

We can now write the previous expression as
\begin{align}
q\left(X\right) & \text{\ensuremath{\propto}}\sum_{n=0}^{\min(n_{1},n_{2})}\sum_{\left\{ \gamma_{1},...,\gamma_{n}\right\} \in\Gamma}\sum_{U^ {}\uplus W=X}q_{2}^{\mathrm{p}}\left(U\right)q_{1}^{\mathrm{p}}\left(U\right)\nonumber \\
 & \times\sum_{X^{1}\uplus...\uplus X^{n_{1}}\uplus Y^{\gamma_{1}^{u,2}}\uplus...\uplus Y^{\gamma_{n_{2}-n}^{u,2}}=W}\nonumber \\
 & \times\prod_{i=1}^{n}q_{1}^{\gamma_{i}(1)}\left(X^{\gamma_{1}(1)}\right)q_{2}^{\gamma_{i}(2)}\left(X^{\gamma_{1}(1)}\right)\nonumber \\
 & \times\prod_{i=1}^{n_{1}-n}\left[q_{1}^{\gamma_{i}^{u,1}}\left(X^{\gamma_{i}^{u,1}}\right)q_{2}^{\mathrm{p}}\left(X^{\gamma_{i}^{u,1}}\right)\right]\nonumber \\
 & \times\prod_{i=1}^{n_{2}-n}\left[q_{1}^{\mathrm{p}}\left(Y^{\gamma_{i}^{u,2}}\right)q_{2}^{\gamma_{i}^{u,2}}\left(Y^{\gamma_{i}^{u,2}}\right)\right]\\
 & =\sum_{U\uplus W=X}q_{1}^{\mathrm{p}}\left(U\right)q_{2}^{\mathrm{p}}\left(U\right)\nonumber \\
 & \times\sum_{n=0}^{\min(n_{1},n_{2})}\sum_{\left\{ \gamma_{1},...,\gamma_{n}\right\} \in\Gamma}\nonumber \\
 & \sum_{X^{1}\uplus...\uplus X^{n_{1}}\uplus Y^{\gamma_{1}^{u,2}}\uplus...\uplus Y^{\gamma_{n_{2}-n}^{u,2}}=W}\nonumber \\
 & \times\prod_{i=1}^{n}q_{1}^{\gamma_{i}(1)}\left(X^{\gamma_{1}(1)}\right)q_{2}^{\gamma_{i}(2)}\left(X^{\gamma_{1}(1)}\right)\nonumber \\
 & \times\prod_{i=1}^{n_{1}-n}\left[q_{1}^{\gamma_{i}^{u,1}}\left(X^{\gamma_{i}^{u,1}}\right)q_{2}^{\mathrm{p}}\left(X^{\gamma_{i}^{u,1}}\right)\right]\nonumber \\
 & \times\prod_{i=1}^{n_{2}-n}\left[q_{1}^{\mathrm{p}}\left(Y^{\gamma_{i}^{u,2}}\right)q_{2}^{\gamma_{i}^{u,2}}\left(Y^{\gamma_{i}^{u,2}}\right)\right].
\end{align}

Let us define 
\begin{align}
q^{\mathrm{p}}\left(U\right) & \propto q_{1}^{\mathrm{p}}\left(U\right)q_{2}^{\mathrm{p}}\left(U\right)\nonumber \\
 & \propto\left[\lambda_{q,1}\right]^{U}\left[\lambda_{q,1}\right]^{U}\nonumber \\
 & =\left[\lambda_{q,1}\cdot\lambda_{q,2}\right]^{U}
\end{align}
 and
\begin{align}
q_{i,j}\left(X\right) & =\frac{q_{1}^{i}\left(X\right)q_{2}^{j}\left(X\right)}{\rho_{i,j}}
\end{align}
where
\begin{align}
\rho_{i,j} & =\int q_{1}^{i}\left(X\right)q_{2}^{j}\left(X\right)\delta X.
\end{align}
Then, the fused density can be written as
\begin{align}
q\left(X\right) & =\sum_{U\uplus W=X}q^{\mathrm{p}}\left(U\right)q^{mbm}\left(W\right)
\end{align}
where $j\in\left\{ 1,...,n_{2}\right\} $
\begin{align}
q^{\mathrm{mbm}}\left(W\right) & \propto\sum_{n=0}^{\min(n_{1},n_{2})}\sum_{\left\{ \gamma_{1},...,\gamma_{n}\right\} \in\Gamma}\nonumber \\
 & \sum_{X^{1}\uplus...\uplus X^{n_{1}}\uplus Y^{\gamma_{1}^{u,2}}\uplus...\uplus Y^{\gamma_{n_{2}-n}^{u,2}}=W}\nonumber \\
 & \times\prod_{i=1}^{n}\rho_{\gamma_{i}(1),\gamma_{i}(2)}q_{\gamma_{i}(1),\gamma_{i}(2)}\left(X^{\gamma_{1}(1)}\right)\nonumber \\
 & \times\prod_{i=1}^{n_{1}-n}\left[\rho_{\gamma_{i}^{u,1},0}q_{\gamma_{i}^{u,1},0}\left(X^{\gamma_{i}^{u,1}}\right)\right]\nonumber \\
 & \times\prod_{i=1}^{n_{2}-n}\left[\rho_{0,\gamma_{i}^{u,2}}q_{0,\gamma_{i}^{u,2}}\left(Y^{\gamma_{i}^{u,2}}\right)\right].
\end{align}

Then, the MBM density is
\begin{align}
 & q^{\mathrm{mbm}}\left(W\right)\nonumber \\
 & \propto\sum_{n=0}^{\min(n_{1},n_{2})}\sum_{\left\{ \gamma_{1},...,\gamma_{n}\right\} \in\Gamma}\left[\prod_{i=1}^{n}\rho_{\gamma_{i}(1),\gamma_{i}(2)}\right]\nonumber \\
 & \times\left[\prod_{i=1}^{n_{1}-n}\rho_{\gamma_{i}^{u,1},0}\right]\left[\prod_{i=1}^{n_{2}-n}\rho_{0,\gamma_{i}^{u,2}}\right]\nonumber \\
 & \times\sum_{X^{1}\uplus...\uplus X^{n_{1}}\uplus Y^{\gamma_{1}^{u,2}}\uplus...\uplus Y^{\gamma_{n_{2}-n}^{u,2}}=W}\prod_{i=1}^{n}q_{\gamma_{i}(1),\gamma_{i}(2)}\left(X^{\gamma_{1}(1)}\right)\nonumber \\
 & \times\prod_{i=1}^{n_{1}-n}q_{\gamma_{i}^{u,1},0}\left(X^{\gamma_{i}^{u,1}}\right)\prod_{i=1}^{n_{2}-n}q_{0,\gamma_{i}^{u,2}}\left(Y^{\gamma_{i}^{u,2}}\right).
\end{align}

This density can also be written as
\begin{align}
 & q^{\mathrm{mbm}}\left(W\right)\nonumber \\
 & \propto\sum_{\gamma\in\Gamma}\left[\prod_{i=1}^{|\gamma|}\rho_{\gamma_{i}(1),\gamma_{i}(2)}\right]\left[\prod_{i=1}^{n_{1}-|\gamma|}\rho_{\gamma_{i}^{u,1},0}\right]\left[\prod_{i=1}^{n_{2}-|\gamma|}\rho_{0,\gamma_{i}^{u,2}}\right]\nonumber \\
 & \times\sum_{X^{1}\uplus...\uplus X^{n_{1}}\uplus Y^{\gamma_{1}^{u,2}}\uplus...\uplus Y^{\gamma_{n_{2}-|\gamma|}^{u,2}}=W}\prod_{i=1}^{|\gamma|}q_{\gamma_{i}(1),\gamma_{i}(2)}\left(X^{\gamma_{1}(1)}\right)\nonumber \\
 & \times\prod_{i=1}^{n_{1}-|\gamma|}q_{\gamma_{i}^{u,1},0}\left(X^{\gamma_{i}^{u,1}}\right)\prod_{i=1}^{n_{2}-|\gamma|}q_{0,\gamma_{i}^{u,2}}\left(Y^{\gamma_{i}^{u,2}}\right).\label{eq:fused_MBM_append}
\end{align}
In the next subsection we calculate the Bernoulli-Bernoulli and Bernoulli-Poisson
fused densities.

\subsection{Calculation of fused densities}

In this section, we calculate the Bernoulli-Bernoulli and Bernoulli-Poisson
fused densities in (\ref{eq:fused_MBM_append}). 

\subsubsection{Bernoulli-Bernoulli densities}

The weight of a Bernoulli-Bernoulli fused density is 
\begin{align}
\rho_{i,j} & =\int q_{1}^{i}\left(X\right)q_{2}^{j}\left(X\right)\delta X\nonumber \\
 & =\left(1-r_{q,1}^{i}\right)\left(1-r_{q,2}^{j}\right)+r_{q,1}^{i}r_{q,2}^{j}\int p_{q,1}^{i}\left(x\right)p_{q,2}^{j}\left(x\right)dx.
\end{align}

We then have that a fused Bernoulli-Bernoulli density is
\begin{align}
q_{i,j}\left(X\right) & =\begin{cases}
r_{i,j}p_{i,j}\left(x\right) & X=\left\{ x\right\} \\
1-r_{i,j} & X=\emptyset\\
0 & \mathrm{otherwise}
\end{cases}
\end{align}
where
\begin{align}
r_{i,j} & =\frac{r_{q,1}^{i}r_{q,2}^{j}\left\langle p_{q,1}^{i},p_{q,2}^{j}\right\rangle }{\rho_{i,j}}\\
p_{i,j}\left(x\right) & =\frac{p_{q,1}^{i}\left(x\right)p_{q,2}^{j}\left(x\right)}{\left\langle p_{q,1}^{i},p_{q,2}^{j}\right\rangle }.
\end{align}

\subsubsection{Bernoulli-Poisson densities}

The weight of a Bernoulli-Poisson fused density is
\begin{align}
\rho_{i,0} & =\int q_{1}^{i}\left(X\right)q_{2}^{\mathrm{p}}\left(X\right)\delta X\nonumber \\
 & =1-r_{q,1}^{i}+r_{q,1}^{i}\left\langle p_{q,1}^{i},\lambda_{q,2}\right\rangle .
\end{align}
Then, a Bernoulli-Poisson fused density is
\begin{align}
q_{i,0}\left(X\right) & =\frac{q_{1}^{i}\left(X\right)q_{2}^{\mathrm{p}}\left(X\right)}{\rho_{i,0}}
\end{align}
where
\begin{align}
q_{i,0}\left(X\right) & =\begin{cases}
r_{i,0}p_{i,0}\left(x\right) & X=\left\{ x\right\} \\
1-r_{i,0} & X=\emptyset
\end{cases}
\end{align}
and
\begin{align}
r_{1,0} & =\frac{r_{q,1}^{i}\left\langle p_{q,1}^{i},\lambda_{q,2}\right\rangle }{\rho_{i,0}}\\
p_{i,0}\left(x\right) & =\frac{p_{q,1}^{i}\left(x\right)\lambda_{q,2}\left(x\right)}{\left\langle p_{q,1}^{i},\lambda_{q,2}\right\rangle }.
\end{align}
This completes the proof of Theorem \ref{thm:Product-PMBs}.

\section{\label{sec:AppendixC}}

In this appendix, we prove Lemma \ref{lem:Minimisation_KLD}. We can
write (\ref{eq:weighted_KLD}) as
\begin{align}
C\left[q\right] & =\omega\int q\left(X\right)\log\frac{q\left(X\right)}{f_{1}\left(X\right)}\delta X\nonumber \\
 & +\left(1-\omega\right)\int q\left(X\right)\log\frac{q\left(X\right)}{f_{2}\left(X\right)}\delta X\nonumber \\
 & =\int q\left(X\right)\log\frac{\left(q\left(X\right)\right)^{\omega}}{\left(f_{1}\left(X\right)\right)^{\omega}}\delta X\nonumber \\
 & +\int q\left(X\right)\log\frac{\left(q\left(X\right)\right)^{1-\omega}}{\left(f_{2}\left(X\right)\right)^{1-\omega}}\delta X.
\end{align}

Now, we apply the upper bound (\ref{eq:upper_bound}) yielding
\begin{align}
 & \log\left(f_{1}\left(X\right)\right)^{\omega}\nonumber \\
 & \leq\log\sum_{X^{0}\uplus...\uplus X^{n_{1}}=X}\left(f_{1}^{\mathrm{p}}\left(X^{0}\right)\right)^{\omega}\prod_{i=1}^{n_{1}}\left(f_{1}^{i}\left(X^{i}\right)\right)^{\omega}.
\end{align}
We apply the similar upper bound to $\log\left(f_{2}\left(X\right)\right)^{1-\omega}$,
which results in the inequality
\begin{align*}
C\left[q\right] & \geq L\left[q\right]
\end{align*}
where 
\begin{align}
L\left[q\right]= & \int q\left(X\right)\log\frac{\left(q\left(X\right)\right)^{\omega}}{\alpha_{1}q_{1}\left(X\right)}\delta X\nonumber \\
 & +\int q\left(X\right)\log\frac{\left(q\left(X\right)\right)^{1-\omega}}{\alpha_{2}q_{2}\left(X\right)}\delta X\nonumber \\
 & =\int q\left(X\right)\log\frac{\left(q\left(X\right)\right)^{\omega}\left(q\left(X\right)\right)^{1-\omega}}{\alpha_{1}q_{1}\left(X\right)\alpha_{2}q_{2}\left(X\right)}\delta X\nonumber \\
 & =\int q\left(X\right)\log\frac{q\left(X\right)}{\alpha_{1}q_{1}\left(X\right)\alpha_{2}q_{2}\left(X\right)}\delta X\nonumber \\
 & =\int q\left(X\right)\log\frac{q\left(X\right)}{\frac{q_{1}\left(X\right)q_{2}\left(X\right)}{\left\langle q_{1},q_{2}\right\rangle }}\delta X-\log\left(\alpha_{1}\alpha_{2}\left\langle q_{1},q_{2}\right\rangle \right).
\end{align}
This completes the proof of Lemma \ref{lem:Minimisation_KLD}. 
\end{document}